\documentclass{article}

\usepackage{arxiv}

\usepackage[utf8]{inputenc} 
\usepackage[T1]{fontenc}    
\usepackage{hyperref}       
\usepackage{url}            
\usepackage{booktabs}       
\usepackage{multicol}
\usepackage{amsfonts}       
\usepackage{nicefrac}       
\usepackage{microtype}      
\usepackage{lipsum}
\usepackage{graphicx}
\usepackage{subcaption}

\usepackage{threeparttable}
\usepackage{siunitx}
\usepackage{booktabs}
\usepackage{hyperref}
\usepackage{bm}
\usepackage{amsmath}
\usepackage{multirow}
\usepackage{tabularx}
\usepackage{array}
\usepackage[ruled,vlined]{algorithm2e}
\usepackage{algpseudocode}
\usepackage{cleveref}
\newtheorem{remark}{Remark}
\usepackage{lineno}
\title{Practical multi-fidelity machine learning: fusion of deterministic and Bayesian models}

\author{
  Jiaxiang Yi \\ Faculty of Mechanical Engineering\\
  Delft University of Technology\\
  Mekelweg 2, Delft, 2628 CD, The Netherlands \\
  \texttt{J.Yi@tudelft.nl} \\
  \And
   Ji Cheng \\
  Department of Computer Science\\
  City University of Hong Kong\\
  83 Tat Chee Avenue, Kowloon, Hong Kong \\
  \texttt{jicheng9617@gmail.com} \\ 
  \And
  Miguel A. Bessa \\
  School of Engineering\\
  Brown University\\
  184 Hope St., Providence, RI 02912, USA \\
  \texttt{miguel\_bessa@brown.edu} \\
}

\begin{document}
\maketitle
\begin{abstract}

Multi-fidelity machine learning methods address the accuracy-efficiency trade-off by integrating scarce, resource-intensive high-fidelity data with abundant but less accurate low-fidelity data. We propose a practical multi-fidelity strategy for problems spanning low- and high-dimensional domains, integrating a non-probabilistic regression model for the low-fidelity with a Bayesian model for the high-fidelity. The models are trained in a staggered scheme, where the low-fidelity model is transfer-learned to the high-fidelity data and a Bayesian model is trained to learn the residual between the data and the transfer-learned model. This three-model strategy -- deterministic low-fidelity, transfer-learning, and Bayesian residual -- leads to a prediction that includes uncertainty quantification for noisy and noiseless multi-fidelity data. The strategy is general and unifies the topic, highlighting the expressivity trade-off between the transfer-learning and Bayesian models (a complex transfer-learning model leads to a simpler Bayesian model, and vice versa). We propose modeling choices for two scenarios, and argue in favor of using a linear transfer-learning model that fuses 1) kernel ridge regression for low-fidelity with Gaussian processes for high-fidelity; or 2) deep neural network for low-fidelity with a Bayesian neural network for high-fidelity. We demonstrate the effectiveness and efficiency of the proposed strategies and contrast them with the state-of-the-art based on various numerical examples and two engineering problems. The results indicate that the proposed approach achieves comparable performance in both mean and uncertainty estimation while significantly reducing training time for machine learning modeling in data-scarce scenarios. Moreover, in data-rich settings, it outperforms other multi-fidelity architectures by effectively mitigating overfitting.

\end{abstract}


\newpage

\section*{List of Abbreviations}
\begin{nolinenumbers}  

\begin{multicols}{2}

\begin{itemize}
    \small
     \item \textbf{DNN} – Deep Neural Network
     \item \textbf{BML} – Bayesian Machine Learning
     \item \textbf{MF} – Multi-Fidelity
    \item \textbf{LF} – Low-Fidelity
    \item \textbf{HF} – High-Fidelity
    \item \textbf{LR} - Linear Regression
    \item \textbf{KRR} – Kernel Ridge Regression
    \item \textbf{GPR} – Gaussian Process Regression
    \item \textbf{KRR-LR-GPR} - The proposed practical multi-fidelity Bayesian machine learning model for the data-scarce scenario, which uses Kernel Ridge Regression as the low-fidelity surrogate, Linear Regression model for transfer-learning, and Gaussian process regression to learn for the residual. 
    \item \textbf{BNN} – Bayesian Neural Network
    \item \textbf{MF-BML} - Multi-Fidelity Bayesian Machine Learning
    \item  \textbf{DNN-LR-BNN} - The proposed practical MF Bayesian machine learning model for the data-rich scenario, which uses Deep Neural Network as the low-fidelity surrogate, Linear Regression model for transfer-learning, and Bayesian Neural Network to learn for the residual. 
    \item \textbf{DNN-BNN} - Multi-fidelity Bayesian Neural Network employing low-fidelity deep neural network and Bayesian neural network as transfer-learning model \cite{Meng2021_MFBNN}. 
    \item \textbf{RBF} – Radial Basis Kernel
    \item \textbf{PCE} – Polynomial Chaos Expansion
    \item \textbf{SGLD} – Stochastic Gradient Langevin Dynamics
    \item \textbf{pSGLD} – preconditioned Stochastic Gradient Langevin Dynamics
    \item \textbf{NRMSE} – Normalized Root Mean Square Error
    \item \textbf{TLL} – Test Log-Likelihood
    \item \textbf{PPD} – Posterior Predictive Distribution
    \item \textbf{QoI} - Quantity of Interest
    \item \textbf{SVE} - Stochastic Volume Element
    \item \textbf{FEA} - Finite Element Analysis
    \item \textbf{SCA} - Self-consistent Clustering Analysis

\end{itemize}
\end{multicols}
\end{nolinenumbers} 

\section{Introduction}


Expressive deterministic (non-probabilistic) machine learning methods, such as DNNs, can approximate any nonlinear continuous function, providing enough data \cite{Hornik1989}. However, they can be prone to over-fitting, especially in the presence of noisy data \cite{Understanding_generalization}, and are not able to predict uncertainty.  BML methods do not have these limitations \cite{Polson2017}, although their training requires additional computational effort. In this article, we present a practical MF strategy that combines non-probabilistic LF surrogates and Bayesian residual surrogates. We demonstrate the simplicity of the proposed MF strategy and consider two common regression scenarios encountered in engineering practice: (1) low-dimensional problems amenable to models with few hyperparameters and easy training; and (2) high-dimensional problems that require more expressive models and involve additional training effort. 

MF data arises naturally in most engineering applications. HF data is often associated with costly and time-consuming experiments, or with accurate yet computationally expensive simulations. In contrast, LF data is usually obtained by efficient strategies (experimental \cite{Chen2022}, analytical \cite{Zhou2021}, or computational \cite{Liu2016, Zhang2021}) at the expense of accuracy.  Therefore, LF data is typically acquired faster, often by several orders of magnitude. Using both datasets can be advantageous to train better models as long as the loss in fidelity is compensated by the presence of more LF data.

The challenges associated with developing MF regression models arise from the need to define one regression model (a surrogate) per fidelity, and in the subsequent definition of the interaction between these models \cite{Toal2015, GiselleFernandezGodino2019}. Without loss of generality, we focus on MF regression models with only two levels of fidelity. We start by introducing MF regression in a general form and by reviewing common choices for surrogate models. We continue by reviewing common MF strategies employing the same type of surrogate model across fidelity levels. Finally, we propose two MF strategies using non-probabilistic LF regression models together with Bayesian residual models.

\textbf{Summary of contributions.} We introduce a practical framework for MF regression and demonstrate its versatility. We hypothesize that most MF regression cases in practice can be covered by considering simple LR transfer-learning models and with one of the following two options for LF and HF models: 1) using KRR and GPR, leading to the KRR-LR-GPR model; or 2) using DNN and BNN, leading to the DNN-LR-BNN model. The KRR-LR-GPR model has few parameters and hyperparameters, facilitating training but limiting scalability (limits on data dimension and size, as well as training and inference time). The DNN-LR-BNN model is more flexible and more challenging to train but is applicable to low- and high-dimensional data with few scalability limits. The simpler model (KRR-LR-GPR) should be used when possible. In limited cases exhibiting complex correlations between LF and HF data, both transfer-learning and residual learning can be achieved simultaneously by a BNN -- denoted as DNN-BNN models \cite{Meng2021_MFBNN}. However, DNN-BNN models are more difficult to train and are expected to have narrower applicability.

\section{Methodology and related work} \label{sec: Retive work}

For simplicity of notation, we start by considering datasets with $d$-dimensional inputs $\mathbf{x}$ but one-dimensional output $y$. Then, the LF dataset contains $\mathbf{x}^l_n$ input and $y^l_n$ output points, where $n=1, ..., N^l$ are the $N^l$ points in this dataset (the superscript $l$ refers to LF). The HF dataset is equivalently defined by $\mathbf{x}^h_n$ and $y^h_n$, where $n=1, ..., N^h$ are the $N^h$ HF points in this dataset. In this article, we propose to define an MF regression model $f^h(\mathbf{x})$ as a combination of three models: (1) the LF model $f^l(\mathbf{x})$ that is trained on LF data $\{\mathbf{x}^l,y^l\}$; (2) a transfer-learning model $g(\mathbf{x})$ that transforms the LF model to the HF data; and (3) a residual model $r(\mathbf{x})$ (if necessary) that captures the difference between transfer-learned LF model and the HF data. Therefore, a general description can be formulated as:
\begin{equation} \label{eq: general_mf_regression}
     f^h(\mathbf{x}) = g\left(f^l(\mathbf{x}), \mathbf{x} \right) + r(\mathbf{x})
\end{equation}

In the data-scarce literature \cite{GiselleFernandezGodino2023,Liu2018_multioutput}, MF models use GPRs for both $f^l(\mathbf{x})$ and $r(\mathbf{x})$ and implicitly assume a linear transfer-learning model $g(\mathbf{x}):= g(f^l(\mathbf{x})) = f^l(\mathbf{x}) \rho$, such that:
\begin{equation} 
\label{eq: linear_mf_formula}
     f^h(\mathbf{x})= f^l(\mathbf{x})\rho + r(\mathbf{x})
\end{equation}
 
\noindent where $\rho$ is a single hyperparameter that transforms the LF model $f^l(\mathbf{x})$ to the HF responses.

Other investigations use more expressive transfer-learning models without defining a different model for the residual \cite{ Meng2021_MFBNN, Cutajar2019}: 
\begin{equation} \label{eq:recurrent_setup}
    f^h(\mathbf{x}) = g\left( f^l(\mathbf{x}), \mathbf{x}\right)
\end{equation}

Here, we argue in favor of generalizing these modeling assumptions by establishing a simple transfer-learning model $g(\mathbf{x})$ that is trained on HF data, choosing an appropriate deterministic LF model $f^l(\mathbf{x})$ and a Bayesian HF model for the residual $r(\mathbf{x})$.

\subsection{Related work for data-scarce scenarios} \label{sec: Data-scarce models}

In the data-scarce regime, GPR or Kriging \cite{Rasmussen2006} stand out due to their elegant derivation and exact integration for Gaussian likelihoods and priors --see \Cref{sec: gpr introduction} for a short introduction. Early in their development, GPR was extended to handle MF problems according to \Cref{eq: linear_mf_formula}, in a method originally called Co-Kriging \cite{Kennedy2000, Forrester2007}. This method is based on expanding the covariance function of GPRs in the form of
\begin{equation*} 
    \mathbf{C} = \begin{bmatrix}
       \mathbf{K} \left( \mathbf{X}^h, \mathbf{X}^h \right) & \mathbf{K} \left( \mathbf{X}^h, \mathbf{X}^l \right) \\
        \mathbf{K} \left( \mathbf{X}^l, \mathbf{X}^h \right) & \mathbf{K} \left( \mathbf{X}^l, \mathbf{X}^l \right) 
    \end{bmatrix}
    \label{eq:kernel_of_MG_GPR}
\end{equation*}

Different variants of this formulation can be found, including multi-task GPR \cite{Liu2018_multioutput, Lin2022a} that consider more than one output, or latent mapping GPR for cases with categorical variables \cite{EweisLabolle2022, Oune2021}. Unfortunately, Co-Kriging has a complexity of order $ \mathcal{O} \left( (N^l+N^h)^3 \right) $ due to the inversion of the covariance matrix $\mathbf{C}$ (see \Cref{eq: concentrated ln-likelihood}), imposing practical limits on the dimensionality and number of training data points that can be considered \cite{Gramacy2020, GiselleFernandezGodino2023, FernandezGodino2019}. In part this issue is mitigated in Hierarchical Kriging \cite{Han2012} and its variants \cite{Han_new_cokriging, Han2013, Lin2022a} by defining a diagonal sub-matrix of $\mathbf{C}$. Recursive Co-Kriging \cite{Gratiet2014_recursive_cokriging} employs a fast hyperparameter optimization strategy and cross-validation to identify the parameter $\rho$, leading to comparable performance to Co-Kriging in some cases. In addition, Scale Kriging \cite{Park2018a_scale, Yi2022} adopts a simplification by considering a fixed parameter with $\rho = 1$ in \Cref{eq: linear_mf_formula} for the transfer-learning model. However, in practice, striking a balance between training accurate MF-GPR models and improving their complexity such that they are trained on large datasets remains a significant challenge \cite{Liu2020, Wu2022}. Furthermore, the literature is scarce on their application to noisy MF data \cite{Giannoukou2024}.

In addition, data-scarce MF problems can be addressed by other methods. For example, MF-PCE \cite{Giannoukou2024} has been applied to noisy HF data and noiseless LF data. This method also relies on \Cref{eq: linear_mf_formula}, but suffers from similar scalability issues as MF-GPR. In contrast, deterministic MF models have shown better scalability but do not predict uncertainty. Examples include MF linear regression \cite{Zhang2018}, MF-RBF regression \cite{Song2019a}, and MF support vector regression \cite{Shi2020}. Inspired by the strengths of deterministic and Bayesian methods, this article explores the combination of deterministic and Bayesian models in MF regression.

\subsection{Related work for data-rich scenarios} \label{sec: Data-rich models}

Machine learning models trained on data-rich MF datasets can be seen in different contexts, including design \cite{Zimmer2021, Liu2019_MFNN_design}, optimization \cite{Li2020_MFNNBO}, and uncertainty quantification \cite{Motamed2020_MF_UQ}. When solving regression problems with deterministic models such as DNNs, predicting the full posterior is avoided by finding a point estimate. For example, the MF-DNN approach proposed by Aydin et al. \cite{Aydin2019} passes data sequentially from the LF to the HF into the same DNN and includes an error metric for fidelity switching. Other strategies involve transfer-learning \cite{Chakraborty2021_transfer, Papez2022} by pre-training the weights of a DNN on the LF dataset and then finalizing training on the HF dataset. Other relevant examples include an MF Graph Neural Network developed by Black et al. \cite{Black2022_MFGNN}, and an MF-DNN architecture introduced by Motamed et al. \cite{Motamed2020_MF_UQ} that establishes the correlation between the HF and LF models via another neural network. Some strategies concatenate LF- and HF-DNN models \cite{Meng2020_MFPINN, Guo2022, Cutajar2019}, as well as Recurrent Neural Networks \cite{Conti2023}, and Convolutional Neural Networks \cite{Wu2023a}. 

Although deterministic MF-DNNs are common in the literature, they are ineffective in predicting uncertainty and are prone to over-fitting \cite{Meng2021_MFBNN, Guo2022, Meng2020_MFPINN}. BNNs \cite{Neal1995} address these limitations but introduce other issues. In particular, the integration of the posterior distribution and the posteriod predictive distribution (PPD) is intractable analytically, so Bayesian inference is performed by Markov Chain Monte Carlo \cite{Neal1995, Welling2011}, Variational Inference \cite{Blundell2015}, Monte Carlo Dropout \cite{Gal2016}, among others \cite{Lakshminarayanan2017}. Unfortunately, Bayesian inference is computationally expensive and makes it challenging to adopt these models for large or high-dimensional datasets. This is visible in the promising work of Meng et al. \cite{Meng2021_MFBNN}, who extended a deterministic MF-DNN architecture \cite{Meng2020_MFPINN} by replacing the HF-DNN with a BNN, concatenating it with a LF-DNN, and using Hamiltonian Monte Carlo to perform inference \cite{Neal2011, Betancourt2018}. Yet, the method is only demonstrated on small datasets. Similar limitations are reported using Bayesian models for both LF and HF data, as in the work of Baptiste et al. \cite{Kerleguer2022_MFBNN} where GPR is used for LF and a BNN for HF. The same authors note that using BNNs for both LF and HF datasets is possible, but the inference time becomes worse than combining GPR with BNN.

The literature spans different architectural choices involving deterministic or Bayesian models. Most strategies rely on the linear model shown in \Cref{eq: linear_mf_formula}, and apply methods to data-scarce scenarios without considering noisy data, especially at the LF level. To our knowledge, addressing the trade-off between inference time and accuracy remains a knowledge gap in the literature.

\section{Practical multi-fidelity Bayesian machine learning} \label{sec: proposed method}

\begin{figure}[h]
    \centering
    \includegraphics[width=0.90\textwidth]{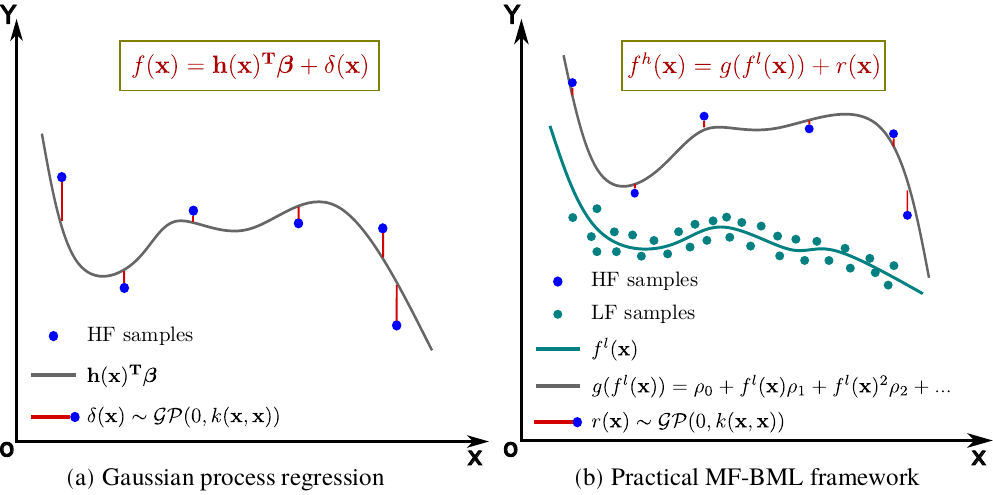}
    \caption{Schematic (a) shows a linear regression model with basis function $\mathbf{h(x)} = [1, \mathbf{x}, \mathbf{x}^2, ...]^T$ and coefficients $\boldsymbol{\beta}$ that is augmented by the residual $\delta(\mathbf{x})$ modeled by zero mean GPR. Schematic (b) shows the proposed MF-BML strategy where the transfer-learning model $g\left(f^l(\mathbf{x}) \right)$ is a linear regression model whose features are the outputs of the LF surrogate model $f^l(\mathbf{x})$ obtained by training on LF data. The linear transfer-learning function that acts on $f^l(\mathbf{x})$ adjusts the LF model to the HF data by determining the coefficients $\boldsymbol{\rho}$, and facilitates the determination of the residual $r(\mathbf{x})$ by a Bayesian model with a simple prior.}
    \label{fig:mf_bml_framework}
\end{figure}

In practice, creating MF models involves so many choices \cite{GiselleFernandezGodino2023, Liu2018_multioutput} that is difficult to perform principled model selection and hyperparameter optimization, especially for large datasets. We propose a simple strategy summarized in \Cref{eq: general_mf_regression} to develop practical MF models based on two ideas, as shown in  \Cref{fig:mf_bml_framework}. First, consider a deterministic LF model $f^l(\mathbf{x})$ together with a probabilistic model of the residual $r(\mathbf{x})$. Second, consider a transfer-learning model $g(\mathbf{x})$ as a simple linear regression model whose features become the outputs of the LF model, i.e. $g\left( f^l(\mathbf{x})\right)$. As we demonstrate in the remainder of the paper, the proposed strategy leverages the advantages of deterministic and probabilistic models while keeping the transfer-learning model simple.

Writing the transfer-learning model $g(\mathbf{x})$ in \Cref{eq: linear_mf_formula} as a linear model (linear in the weights) and considering its features as the LF model leads to:
\begin{align}
  f^h(\mathbf{x}) &=   g\left(f^l(\mathbf{x})\right) + r(\mathbf{x})  \notag \\
   &= \bm{m}(\mathbf{x})^T\bm{\rho}+  r(\mathbf{x})  \notag \\
    &= \rho_0 + f^l(\mathbf{x})\rho_1 + f^l(\mathbf{x})^2\rho_2+ \ldots + f^l(\mathbf{x})^{M-1}\rho_{M-1}+r(\mathbf{x}) \label{eq: mf-bml}
\end{align}
\noindent where $\mathbf{m(x)} = [1, f^l(\mathbf{x}), f^l(\mathbf{x})^2, \ldots, f^l(\mathbf{x})^{M-1}]^T$ is the vector with $M$ polynomial basis features of the LF model, and $\bm{\rho}=[\rho_0, \rho_1, \ldots, \rho_{M-1}]^T$ is the corresponding coefficient vector to be determined from training on the HF data, i.e. these coefficients are not imposed as hyperparameters. This contrasts with the models reviewed in \Cref{sec: Data-scarce models} that reduce \Cref{eq: mf-bml} to \Cref{eq: linear_mf_formula} by considering $\mathbf{m(x)} = f^l(\mathbf{x})$ and then considering $\rho_1$ as a hyperparameter.

In addition to defining the transfer-learning model, there are many possible choices for LF and HF models. \Cref{tab: configurations of MF-BML} summarizes common models, as reviewed in the introduction. We suggest two different MF model choices for spanning a large number of data-scarce and data-rich MF regression problems: 1) kernel ridge regression together with Gaussian process regression (KRR-LR-GPR)\footnote{Naming convention for the MF model (<LF model>-<transfer-learning model>-<residual model>): 
 the abbreviation on the left is the surrogate for the LF model,  the middle is the transfer-learning model, and the last abbreviation is the model for the residual. In the absence of one of the models, for example not considering the residual, then the abbreviation only has two models (e.g. DNN-BNN refers to a DNN model for LF with a BNN doing the transfer-learning and no residual).} for data-scarce or low-dimensionality scenarios; and 2) DNN together with a Bayesian neural network (DNN-LR-BNN) for data-rich or multi-output scenarios.
\begin{center}
    \renewcommand{\arraystretch}{1.2} 
    \begin{table}[h]
        \centering
        \caption{Candidates for LF and HF surrogates within the MF-BML framework}
        \label{tab: configurations of MF-BML}
        \begin{threeparttable}[c]
            \begin{tabular}{m{6cm} m{6cm}}
                \hline
                LF surrogates                        & HF surrogates                      \\
                \hline
                Linear regression (LR)                  & Gaussian process regression (GPR)         \\
                Kernel ridge regression (KRR)                   &  Polynomial chaos expansion (PCE)  \\
                Deep neural network (DNN)                              & Bayesian neural network (BNN)                \\
                Support Vector Regression (SVM)           &  ...                              \\
                Gaussian process regression (GPR) \tnote{*} &                                    \\
                Polynomial chaos expansion (PCE) \tnote{*}                                  &                                    \\
                ...                                  &                                    \\
                \hline
            \end{tabular}
            \begin{tablenotes}
                \item [*] \small Often used in conjunction with the same model at the HF, and considering $\mathbf{m(x)} = f^l(\mathbf{x})$. Examples: GPR-LR-GPR in the form of Co-Kriging \cite{Kennedy2000} or Hierarchical Kriging \cite{Han2012}; PCE-LR-PCE \cite{Giannoukou2024}; among others \cite{Park2018_including}. 
            \end{tablenotes}
        \end{threeparttable}
    \end{table}
\end{center}


\begin{remark} \label{remark 1}
    LF datasets are usually larger than HF datasets because, by definition, LF data is acquired faster than HF data. If the LF dataset is large, deterministic LF models are advantageous because probabilistic models are not sufficiently scalable. In addition, deterministic LF models have lower complexity and fewer hyperparameters, while still being able to handle noisy data (aleatoric uncertainty). 
\end{remark}

\begin{remark} \label{remark 2}
    The key disadvantage of choosing deterministic LF models is their inability to characterize epistemic (or model) uncertainty (at the LF). However, this will be shown to have limited relevance for the LF model when the residual or the transfer-learning model is probabilistic. In such cases, the HF predictions include uncertainty quantification and harness the typical advantages of probabilistic models.
\end{remark}

\subsection{Model for data-scarce or low-dimensional scenarios: KRR-LR-GPR}

GPRs are one of the most successful models for data-scarce scenarios \cite{Rasmussen2006}, having only a few hyperparameters and performing Bayesian inference for Gaussian observation distributions and priors without needing numerical integration. Consequently, they are easy to train for small datasets, making them an important HF model.

We argue that a logical LF model to pair with GPRs in a data-scarce HF scenario is KRR, i.e. the deterministic formulation of GPR \cite{Forrester2007}. KRR is a kernel machine learning method that is a point estimate of a GPR with the same kernel, therefore also equivalent to a DNN with an infinitely wide hidden layer \cite{Forrester2008}. KRR is robust to noisy data, and easy to train due to having few hyperparameters. More importantly, existing KRR approaches can lower complexity to a range between $\mathcal{O}(N^2)$ and $\mathcal{O}(N)$, thus handling larger datasets than PCE or GPR \cite{Shabat2021, Meanti2020} -- invaluable for LF datasets because they are typically larger than HF ones. Other LF models such as LR or DNNs (e.g., see Table \ref{tab: configurations of MF-BML}) have important drawbacks. LR models are sensitive to the choice of the basis functions (often considered to be polynomials whose order is a hyperparameter); this is why we argue for their use as a simple transfer-learning model $g(\mathbf{x})$ but not as a LF or residual model. Conversely, DNNs have a large number of hyperparameters and are therefore less practical to train. Good practice involves starting with simpler models such as KRR that are easier to train and only considering DNNs when the simpler methods fail. We note that GPRs can also be a good choice for a LF model if the LF dataset is small, but training and inference time increases when compared to KRR, as shown later. 

A common kernel of choice for both KRR and GPR when there is not enough prior information is the RBF kernel \cite{Forrester2007}:
\begin{equation} \label{eq:rbf kernel}
    k(\mathbf{x}^{i}, \mathbf{x}^{j}) = \exp \left(-\sum_{c=1}^{d} \theta_l\left({x}_{c}^{i}-{x}_{c}^{j}\right)^2\right)
\end{equation}

\noindent where $\bm{\theta}$ is a $d$-dimensional vector that controls the length scale of each dimension.

The proposed KRR-LR-GPR model established on \Cref{eq: mf-bml} is easy to implement and the main steps are listed in  Algorithm \ref{alg: KRR-LR-GPR}, while additional details are provided in \Cref{sec: paramete_estimation_of_mf_rbf_gp}. Recall that the key idea is to train a GPR model where, instead of considering a zero mean function, we assume it to be a linear model whose features are the LF model. This is advantageous as there is an explicit solution for finding the parameters $\boldsymbol{\rho}$ from the HF data, no longer treating them as hyperparameters: Step 2.1 in Algorithm \ref{alg: KRR-LR-GPR} or \Cref{eq: estimation_of_rho} in \Cref{sec: paramete_estimation_of_mf_rbf_gp}. The mean model of the GPR then captures most of the HF response by linear transfer-learning of the LF model, leaving the remaining nonparametric approximation for the residual via a Gaussian process whose kernel hyperparameters are optimized as usual (Step 2.2 in Algorithm \ref{alg: KRR-LR-GPR}). The resulting KRR-LR-GPR model predicts the response and corresponding uncertainty, usually by following a Normal distribution $\mathcal{N}(\hat{f}^h(\mathbf{x}), \hat{\sigma}_{h}^2(\mathbf{x}))$ to ensure fast training and inference (exact integration of posterior and posterior predictive distributions). We find that a first-order linear transfer-learning model including a bias term, i.e. $\mathbf{m}(\mathbf{x}) = [1,  f^l(\mathbf{x})]^T$, provides robust results according to the ablation study for different problems (see  \Cref{sec: lf basis selection}).

\begin{algorithm}[H]
    \label{alg: KRR-LR-GPR}
    \SetAlgoLined
    \KwData{LF dataset $\mathcal{D}(\mathbf{X}^l, \mathbf{y}^l)$, HF dataset $\mathcal{D}(\mathbf{X}^h, \mathbf{y}^h)$}
    \KwResult{$\mathcal{N}(\hat{f}^h(\mathbf{x}), \hat{\sigma}_{h}^2(\mathbf{x}))$}

     \textbf{Step 1}: Train $f^l(\mathbf{x})$ by optimizing $\bm{\theta}^{l}$ based on LF dataset $\mathcal{D}(\mathbf{X}^l, \mathbf{y}^l)$  \\
   
    \textbf{Step 2}: Train $g(\mathbf{x})$ and  $r(\mathbf{x})$ based on  $\mathcal{D}(\mathbf{X}^h, \mathbf{y}^h)$ and  $f^l(\mathbf{x})$ \\
    
    $\quad$\textbf{Step 2.1}: Calculate $\hat{\bm{\rho}} = \left (\mathbf{m}(\mathbf{X}^h)\mathbf{K}(\mathbf{X}^h,\mathbf{X}^h )^{-1}\mathbf{m}(\mathbf{X}^h)^T \right)^{-1} \mathbf{m}(\mathbf{X}^h)\mathbf{K}(\mathbf{X}^h,\mathbf{X}^h )^{-1}\mathbf{y}^h$  \\ 
    $\quad$\textbf{Step 2.2}: Optimize concentrated ln-likelihood function for $\bm{\theta}^{h}$  \\ 
    
    \caption{KRR-LR-GPR model training}
  \end{algorithm}

\subsection{Model for data-rich or high-dimensional scenarios: DNN-LR-BNN} \label{sec: MF-BNN-DNN architecture}

\begin{figure}[h]
    \centering
    \includegraphics[width=\textwidth]{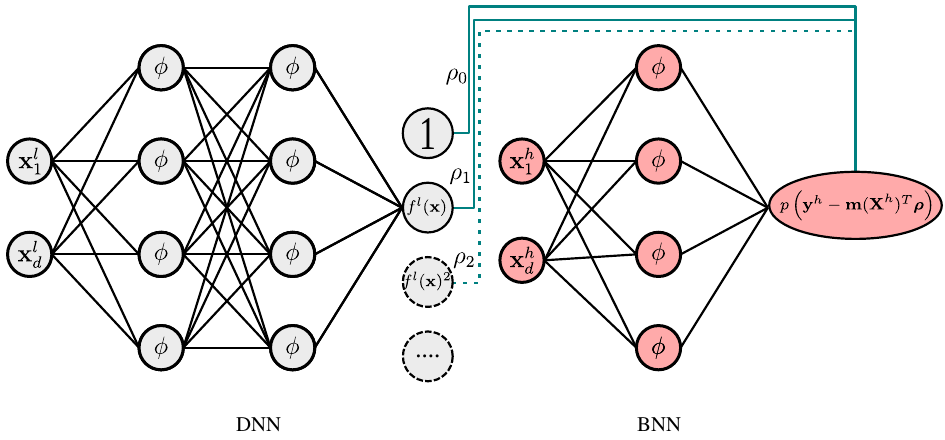}
    \caption{Schematic of the DNN-LR-BNN architecture. The DNN is trained on LF data, then it is used as a basis function of a linear transfer-learning model $g$ that better explains the HF data and that is represented by the green connections, leading to $\mathbf{m}(\mathbf{x}^h)^T\boldsymbol{\rho}$. The BNN is then trained on the residual $\mathbf{r}(\mathbf{x}^h)=\mathbf{y}^h-\mathbf{m}(\mathbf{x}^h)^T\boldsymbol{\rho}$.}
    \label{fig: DNN-LR-BNN}
\end{figure}

Pairing a DNN as the LF model with a BNN in a MF model leverages the scalability of DNNs with the predictive abilities of BNNs. We propose a novel configuration summarized in \Cref{fig: DNN-LR-BNN} labeled DNN-LR-BNN, where transfer-learning is done via LR as in \Cref{eq: mf-bml}, and the residual is modeled by a BNN. 

In the DNN-LR-BNN model, the BNN learns the residual $r(\mathbf{x}^h) = \mathbf{y}^h - \mathbf{m}(\mathbf{x}^h)^T\bm{\rho}$, where the order of $\mathbf{m}(\mathbf{x}^h)$ is a hyperparameter. Usually, we assume Normal distributions for the prior and observation distribution of BNNs, leading to a Normal PPD; Therefore, the DNN-LR-BNN model also has a Normal PPD. As mentioned previously, there are many viable strategies for inference in BNNs: from Markov Chain Monte Carlo approaches like Hamiltonian Monte Carlo \cite{Neal2011, Betancourt2018} to SGLD \cite{Welling2011}, passing through Variational Inference approaches such as Bayes by Back-propagation \cite{Blundell2015}. In practice, for large datasets, we recommend the use of a variant of SGLD called pSGLD \cite{Li2015} due to its superior scalability \cite{Deng2022} -- additional details are provided in \Cref{sec:pSGLD}. 

Contrary to the KRR-LR-GPR model, the determination of the transfer-learning parameters $\bm{\rho}$ is not done by  \Cref{eq: estimation_of_rho} because it becomes intractable to calculate the covariance matrix of a BNN. Simultaneously, using cross-validation or other hyperparameter tuning strategies can be computationally intensive due to the time needed for BNN inference. Therefore, we propose to estimate $\bm{\rho}$ by solving the following optimization problem before training the BNN:
\begin{equation}
\label{eq: obj of hyperparameter}
\hat{\bm{\rho}} = \arg \min_{\bm{\rho}} \sum_{i=1}^{N^h} \left\| \mathbf{y}^h - \mathbf{m}(\mathbf{X}^h)^T \bm{\rho} \right\|^2
\end{equation}

This equation is easy to optimize, akin to what is done in linear regression. Note that if we assumed $\rho_0 =0$ and $\rho_1 =1$, a vanilla MF model would be obtained (fusing a DNN with a BNN without additional parameters). The main steps of DNN-LR-BNN are listed in Algorithm \ref{alg: DNN-LR-BNN}, leading to a prediction following a Normal distribution $\mathcal{N}(\hat{f}^h(\mathbf{x}), \hat{\sigma}_{h}^2(\mathbf{x}))$ at any unknown point.

\begin{algorithm}[H]
    \label{alg: DNN-LR-BNN}
    \SetAlgoLined
    \KwData{LF dataset $\mathcal{D}(\mathbf{X}^l, \mathbf{y}^l)$, HF dataset $\mathcal{D}(\mathbf{X}^h, \mathbf{y}^h)$}
    \KwResult{$\mathcal{N}(\hat{f}^h(\mathbf{x}), \hat{\sigma}_{h}^2(\mathbf{x}))$}
    \textbf{Step 1}: Train DNN based on LF dataset $\mathcal{D}(\mathbf{X}^l, \mathbf{y}^l)$ and corresponding DNN settings \\ 
    \textbf{Step 2}: Train $g(\mathbf{x})$ and  $r(\mathbf{x})$ based on  $\mathcal{D}(\mathbf{X}^h, \mathbf{y}^h)$ and  $f^l(\mathbf{x})$ \\
    $\quad$\textbf{Step 2.1}: Obtain  $\bm{\rho}$ by minimizing \Cref{eq: obj of hyperparameter}\\ 
    $\quad$\textbf{Step 2.2}: BNN inference with pSGLD\\
    \caption{DNN-LR-BNN model training}
  \end{algorithm}

\section{Experiments} \label{sec: experiments}

We separately analyze the performance of KRR-LR-GPR and DNN-LR-BNN models and compare them with state-of-the-art strategies. For comparison with the KRR-LR-GPR method, we select Co-Kriging \cite{Forrester2007}, Hierarchical Kriging \cite{Han2012}, and Scaled Kriging \cite{Park2018a_scale}. We first consider 10 different functions to be learned using datasets with two fidelity levels, as described in \Cref{sec: low dimensional functions}. These are low-dimensional functions with an input dimension spanning from $d=1$ to $d=8$, so they only need a small number of HF training samples. Importantly, the datasets utilized in this section are assumed to be noisy\footnote{We implemented the existing methods such that noisy datasets could also be considered.}. Additional experiments are conducted in  \Cref{sec:additional experiments for KRR-LR-GPR} where noiseless datasets are also considered. Then, we train KRR-LR-GPR to predict the aerodynamic coefficients for a NACA 0012 airfoil in Section \ref{sec:naca0012_airfoil}. Concerning the performance assessment of DNN-LR-BNN, comparison with other models is more challenging due to the multitude of hyperparameters and model choices that are possible when considering neural networks in different fidelity levels. Therefore, we compare DNN-LR-BNN with single-fidelity BNN and DNN-BNN models \cite{Meng2021_MFBNN} considering the same hyperparameters, as summarized in  \Cref{sec: hyperparameters of DNN-LR-BNN}. Several different numerical examples are considered in \Cref{sec: 1d_mf_dnn_bnn} and \Cref{sec:dnn_lr_bnn_higher_dimensional_examples}; and we finally validate its performance with a material structure-property linkages problem in Section \ref{sec:mat_law_prediction}.

For model comparison, considering that no single metric can objectively evaluate model performance, we use NRMSE and R2 Score to assess the predicted mean performance \cite{Lee2024}.  Concerning the uncertainty quantification metric, we adopt the TLL when we only have access to the noisy test dataset, as occurs in practice \cite{Deshpande2024}. In addition, we implemented all the algorithms by ourselves and in the same coding language to record the CPU execution time for LF and HF. Each experiment is conducted on a node of an HPC cluster platform with an Intel® Xeon(R) E5-2643v3 CPU with 6 cores of 3.40GHz and 128 GB of RAM. Rigorous comparison of computational cost is challenging, as each method can be optimized differently. Our implementations and results are made publicly available, in an attempt to demonstrate fairness in our comparisons and motivate future research on this topic.

\subsection{Experiments with data-scarce HF and low-dimensional problems: KRR-LR-GPR} \label{sec: 1d noisy illustrative examples}

\subsubsection{Illustrative example} 

\begin{figure}[h]
    \centering
    \includegraphics[width=\textwidth]{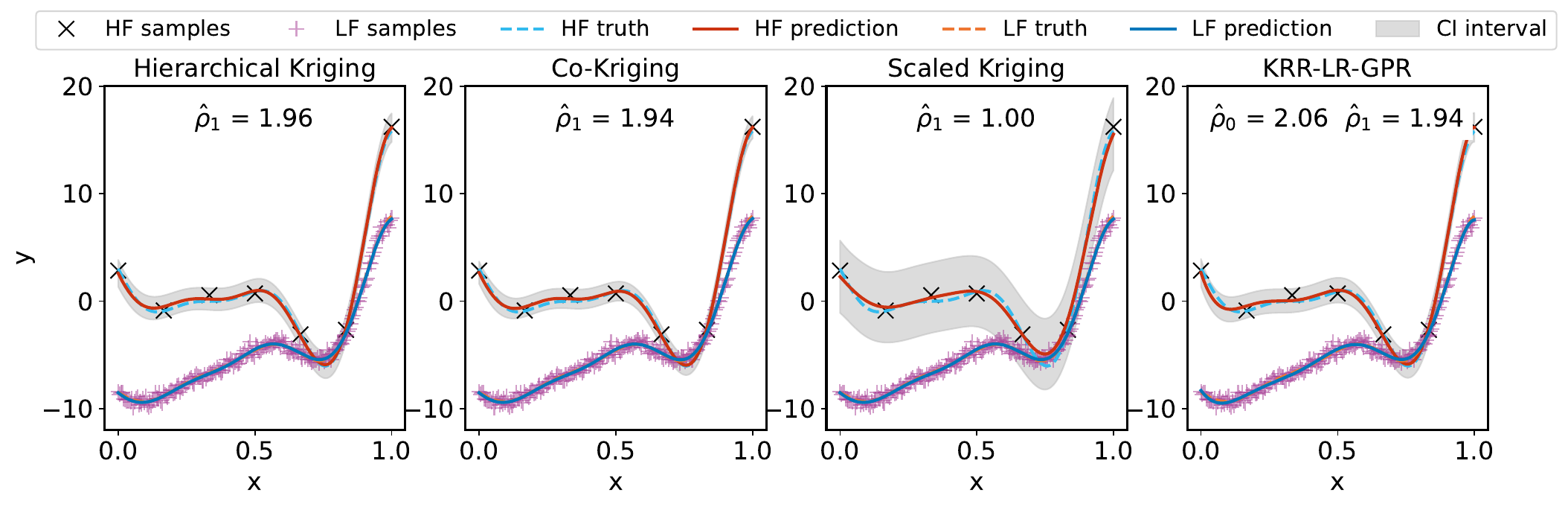}
    \caption{Fitting performance of KRR-LR-GPR  on the illustrative example. }
    \label{fig:1d_noisy_gpr_illustration_case}
\end{figure}

\begin{figure}[h]
    \centering
    \includegraphics[width = \textwidth]{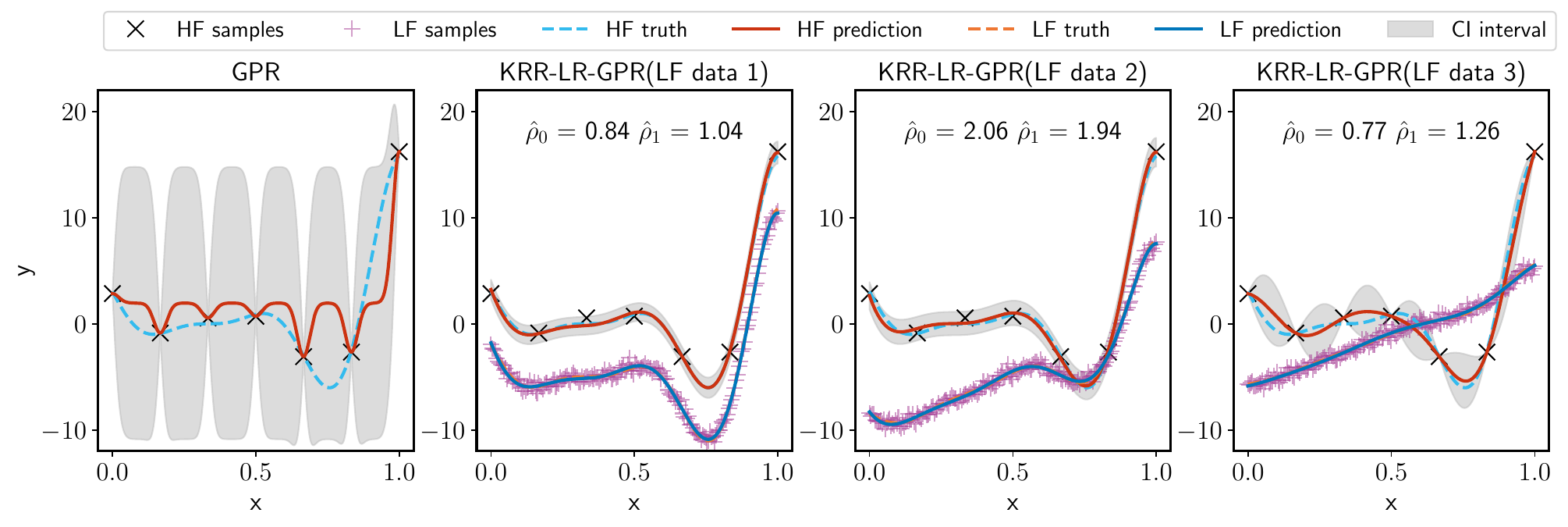}
    \caption{Predictions of KRR-LR-GPR for the Forrester function with different LF datasets of decreasing correlation with the HF dataset (from left to right), as indicated in \Cref{tab:1d results with noise by mfrbfgpr}. Different correlations between LF and HF data are obtained according to \Cref{eq: forrester function2}, where the parameters $A$, $B$, and $C$ of the LF function are changed as follows: \textit{LF data 1} considers $A=1, B=0, C=5$; \textit{LF data 2} considers $A=0.5, B=10, C=5$; \textit{LF data 3} considers $A=0.1, B=10, C=0.1$. The figure on the left corresponds to the single-fidelity result of GPR.}
    \label{fig:1d illustrative example of mf_rbf_gpr}
\end{figure}

\begin{figure}[h]
    \centering
    \includegraphics[width=\textwidth]{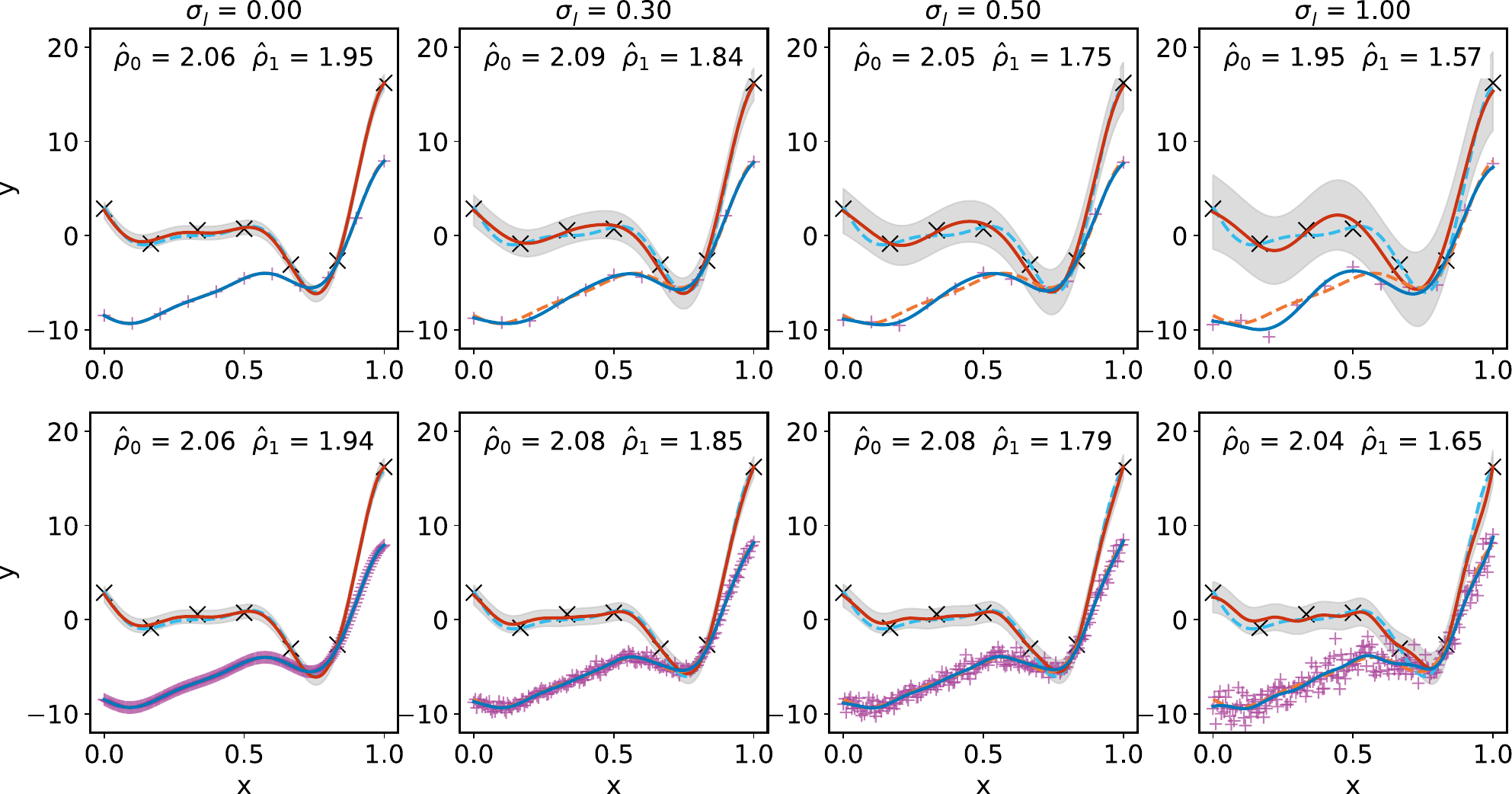}
    \caption{Performance of KRR-LR-GPR model on Forrester function with different number of LF samples and noise levels: the top and bottom rows show results obtained for a LF dataset with 11 and 200 samples, respectively. In this figure, the HF dataset is not changed and considers 7 samples. The noise level of LF is indicated by $\sigma_l$. The number of LF samples is clear from the cross markers in each plot.}
    \label{fig: MF_RBF_GPR_different_noise_influence}
\end{figure}

Given the traditional importance of the Forrester function \cite{Forrester2007, Han2012}, we start by illustrating the effectiveness of the KRR-LR-GPR method for this example -- see \Cref{eq: forrester function}. First, we compare the proposed method with methods in the literature that involve GPR models for both fidelity levels.  We begin in  \Cref{fig:1d_noisy_gpr_illustration_case} and \Cref{fig:1d illustrative example of mf_rbf_gpr} by considering LF and HF datasets with 200 and 7 uniformly sampled points, respectively. Note that our datasets include Gaussian noise $\mathcal{N}(0, 0.3^2)$ on both fidelity levels (LF and HF). \Cref{fig: MF_RBF_GPR_different_noise_influence} considers different LF dataset sizes and different noise levels. Unsurprisingly, due to the simplicity of the Forrester function and its low-dimensionality, this first example was found to be trivial for most methods.

\begin{table}[hbt!]
    \centering
    \caption{Performance comparison between KRR-LR-GPR and other MF-GPR methods on the illustrative example over 5 independent runs.}
    \label{tab:results_1D_noisy_mfrbfgpr}
    \begin{tabular}{c m{1.5cm} m{1.5cm} m{1.5cm}m{1.5cm} m{1.5cm} m{1.5cm}}
        \hline
        \multirow{2}{*}{Methods}    &  \multirow{2}{*}{NRMSE}    &   \multirow{2}{*}{R2 Score}         &  \multirow{2}{*}{TLL}  &  \multirow{2}{*}{$\hat{\sigma}_h$}   &  \multicolumn{2}{c}{Training time($s$)} \\
       & & & & & LF  & HF \\
        \hline
        Hierarchical Kriging          &  \textbf{0.1408} ($\pm 0.044$) & \textbf{0.9927} ($\pm 0.005$)      &  \textbf{-0.7895} ($\pm 0.189$)          &  0.4360 ($\pm 0.167$) & 4.2705 ($\pm 1.048$) & 0.2429 ($\pm 0.056$) \\
        Co-Kriging     & 0.1429 ($\pm 0.041$)    & 0.9925 ($\pm 0.004$)       & -0.9651 ($\pm 0.468$)         & \textbf{0.3446} ($\pm 0.178$)  & 4.2305 ($\pm 0.676$)& 0.4458 ($\pm 0.078$)       \\
        Scaled Kriging    & 0.2435 ($\pm 0.032$)       & 0.9796 ($\pm 0.005$)       & -1.3306 ($\pm 0.359$)        &0.5284 ($\pm 0.485$)    & 4.0597 ($\pm 0.792$) & 0.2202 ($\pm 0.038$)   \\
        KRR-LR-GPR   &  0.1455 ($\pm 0.047$) &0.9922 ($\pm 0.005 $)&  -0.8573 ($\pm 0.308$) & 0.3803 ($\pm 0.164$) & \textbf{2.4128 ($\pm 1.348$)} & 0.2422 ($\pm 0.037$) \\
        \hline
    \end{tabular}
\end{table}

\Cref{fig:1d_noisy_gpr_illustration_case} shows the results for all three MF-GPR models compared to the proposed KRR-LR-GPR model (for one of the 5 realizations chosen at random). \Cref{tab:results_1D_noisy_mfrbfgpr} summarizes the performance metrics obtained from 5 independent runs with random seeds for each method. As mentioned, the Forrester function is easily predicted by all methods, except Scaled Kriging. The HF predicted means overlap almost perfectly with the HF ground-truth, as quantified in  \Cref{tab:results_1D_noisy_mfrbfgpr} by an NRMSE approaching zero and an R2 Score close to 1. Co-Kriging, Hierarchical Kriging, and KRR-LR-GPR all converge to similar $\hat{\rho}_1$. Note that Scaled Kriging considers $\hat{\rho}_1 = 1$, leading to inferior predictions. Unsurprisingly, KRR-LR-GPR requires less training time than all MF-GPR methods due to the use of KRR for the LF.
 
\Cref{fig:1d illustrative example of mf_rbf_gpr} and \Cref{tab:1d results with noise by mfrbfgpr} summarize the performance of KRR-LR-GPR when considering datasets with different correlations between LF and HF data. Three different LF functions are used to generate three different LF datasets, as explained in \Cref{sec: low dimensional functions}, leading to the labels \textit{LF data 1}, \textit{LF data 2} and \textit{LF data 3} in \Cref{fig:1d illustrative example of mf_rbf_gpr}.

\begin{table}[h]
    \centering
    \caption{Summary of performance of KRR-LR-GPR model considering different LF datasets over 5 independent runs.}
    \label{tab:1d results with noise by mfrbfgpr}
    \renewcommand{\arraystretch}{1.2} 
    \begin{tabular}{m{2.5cm} m{1cm} m{2.5cm} m{2.5cm} m{2.5cm}m{2.5cm}}
        \toprule
         \multirow{2}{*}{Methods} & \multirow{2}{*}{} & GPR & \multicolumn{3}{c}{KRR-LR-GPR} \\ 
         & &  & LF data 1 & LF data 2 & LF data 3 \\
        \midrule
        \multicolumn{2}{l}{Pearson cor. coef. $r$} & --&  1.0    & 0.737  & 0.407 \\
        \multicolumn{2}{l}{NRMSE}                  & 1.3176 ($\pm 0.006$)   & 0.0823 ($\pm 0.024$) & 0.1216 ($\pm 0.025$) & 0.6524 ($\pm 0.057$) \\
        \multicolumn{2}{l}{R2 Score}               & 0.4147 ($\pm 0.005$)  & 0.9975 ($\pm 0.001$) & 0.9948 ($\pm 0.002$) & 0.8556 ($\pm 0.024$)\\
        \multicolumn{2}{l}{TLL}                    & -3.7502 ($\pm 1.490$) & -0.5303 ($\pm 0.300$) & -0.8650 ($\pm 0.391$) & -2.2474 ($\pm 0.094$) \\
        \midrule
        \multirow{2}{*}{Training time ($s$)} & HF  & 0.1560 ($\pm 0.031$)  & 0.1731($\pm 0.023$) & 0.1707 ($\pm 0.016$)& 0.1398 ($\pm 0.039$) \\  
                                             & LF  & --       & 1.5543 ($\pm 0.562$)& 1.3730 ($\pm 0.326$) & 1.2043 ($\pm 0.523$) \\
        \midrule
        \multicolumn{2}{l}{$\hat{\sigma}_{h}$} & 0.000060 ($\pm 0.000001$) & 0.2851 ($\pm 0.050$) & 0.2447 ($\pm 0.082$) & 2.6200 ($\pm 0.307$)\\
        \bottomrule
    \end{tabular}
\end{table}

\begin{figure}[h]
    \centering
    \includegraphics[width=\textwidth]{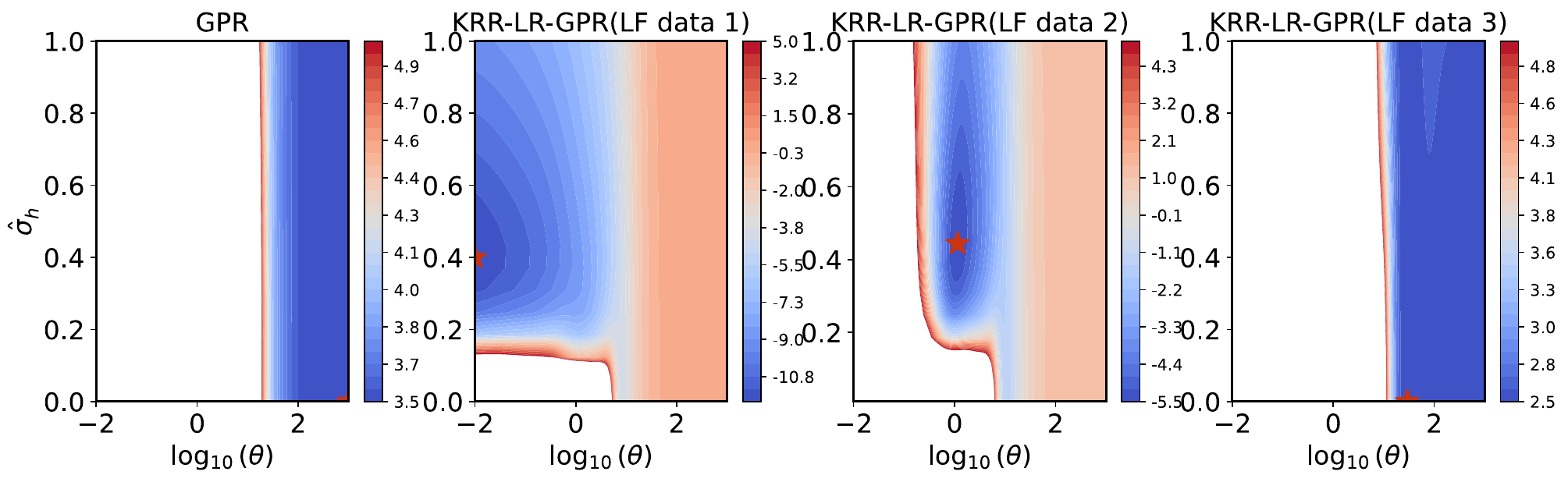}
    \caption{Negative log marginal likelihood values of GPR and KRR-LR-GPR within parameter space. The contour plots only show regions where the negative log marginal likelihood value is smaller than 5 for better illustration, the maximum values for each method are more than $10^7$. The X-axis represents kernel parameter $\log_{10}(\theta)$ and the Y-axis is the estimated noise level $\hat{\sigma}_h$, the red point is the best parameter found for every method by "L-BFGS-B" with 10 restarts.}
    \label{fig: negative log likelihood}
\end{figure}

\Cref{fig:1d illustrative example of mf_rbf_gpr} and \Cref{tab:1d results with noise by mfrbfgpr} clarify that the single-fidelity GPR model cannot learn the underlying function well because it fails to identify the noise level. However, the transfer-learned LF model of KRR-LR-GPR provides appropriate conditioning of the GPR residual and leads to good predictions (see appropriate performance scores, NRMSE, and R2 Score). Performance deteriorates as the Pearson correlation coefficient $r$ between the two fidelity levels decreases from \textit{LF data 1} to \textit{3}. Similarly, model uncertainty at the HF is lower when the LF datasets are more correlated with the HF dataset (higher Pearson correlation coefficient). \Cref{fig: negative log likelihood} also explains this by showing the negative log marginal likelihood landscape in the parameter space for a particular run of each of the three cases, where it is highlighted that in the third case (\textit{LF data 3}) there is no clear optimum value (rightmost figure), exhibiting a similar marginal likelihood to the single-fidelity GPR model (leftmost figure) -- this explains why the noise level at the HF is not properly estimated. The results are intuitive: if the LF data has limited or nonexistent correlation with the HF, then there is a degradation of the prediction quality of the MF model.

\begin{table}[h]
    \centering
    \caption{Results of KRR-LR-GPR on Forrester function with different LF samples and LF noise levels over 5 independent runs}
    \label{tab: MF_RBF_GPR_different_noise_influence}
    \renewcommand{\arraystretch}{1.2}
    \begin{tabular}{c c m{2.5cm} m{2.5cm} m{3.5cm}m{2.5cm}}
        \hline
        LF samples & $\sigma_{l} $       & NRMSE  & R2 Score &  TLL & $\hat{\sigma}_{h}$\\
        \hline
      \multirow{4}{2em}{11}   &  0.0& 0.1267 ($\pm 0.012$) &0.9945 ($\pm 0.001$)    & -0.6279 ($\pm 0.007$)& 0.3300 ($\pm 0.052$)        \\
        & 0.3  & 0.2572 ($\pm 0.088$) & 0.9751  ($\pm 0.018$)   &  -11304.37 ($\pm 14787.37$)& 0.3229 ($\pm 0.378$)       \\
        & 0.5 & 0.3184 ($\pm 0.073$) & 0.9640  ($\pm 0.016$)   &-11202.4 ($\pm 14779.75$) & 0.4992 ($\pm 0.523$)      \\
          & 1.0   & 0.4125 ($\pm 0.139$)  & 0.9361 ($\pm 0.045$)     &  -11128.88 ($\pm 14842.75$) & 0.6761  ($\pm 0.682$) \\ 
        \hline
         \multirow{4}{2em}{200}   &  0.0  &0.1184 ($\pm 0.015 $) &0.9952 ($\pm 0.001$)     &-0.6106 ($\pm 0.092$) & 0.3374 ($\pm 0.057$)       \\
        & 0.3& 0.1119 ($\pm 0.016$) & 0.9957 ($\pm 0.001$)  &  -0.5398 ($\pm 0.080$) &0.3159 ($\pm 0.017$)      \\
        &0.5  & 0.1184 ($\pm 0.020$) & 0.9951 ($\pm 0.001$)    &  -0.5693 ($\pm 0.094$)& 0.3136 ($\pm 0.061$)       \\
          & 1.0 &0.1563 ($\pm 0.026$) & 0.9915 ($\pm 0.003$)     & -0.7243 ($\pm 0.121$) & 0.3731 ($\pm 0.055$)      \\ 
        \hline
    \end{tabular}
\end{table}

Furthermore, we investigate the performance of the KRR-LR-GPR model with different noise levels and different LF dataset sizes (see \Cref{fig: MF_RBF_GPR_different_noise_influence} and \Cref{tab: MF_RBF_GPR_different_noise_influence}).  We consider different Gaussian noise for the LF with standard deviations of $\sigma_{l}=0.0$, $\sigma_{l}=0.3$, $\sigma_{l}=0.5$, and $\sigma_{l}=1.0$, such that the robustness of KRR-LR-GPR to LF noise is assessed while maintaining the HF noise as $\sigma_{h}=0.3$. The figure and table clarify that the model predicts both the function and noise level accurately for the cases with enough LF data. Conversely, it struggles to identify proper noise levels when the LF dataset only has 11 points and for higher LF noise levels, as expected.

\subsubsection{Comprehensive numerical experiments of KRR-LR-GPR} \label{sec: comprehensive experiments of KRR-LR-GPR}

\begin{figure}[hbt!]
    \centering
    \includegraphics[width=\textwidth]{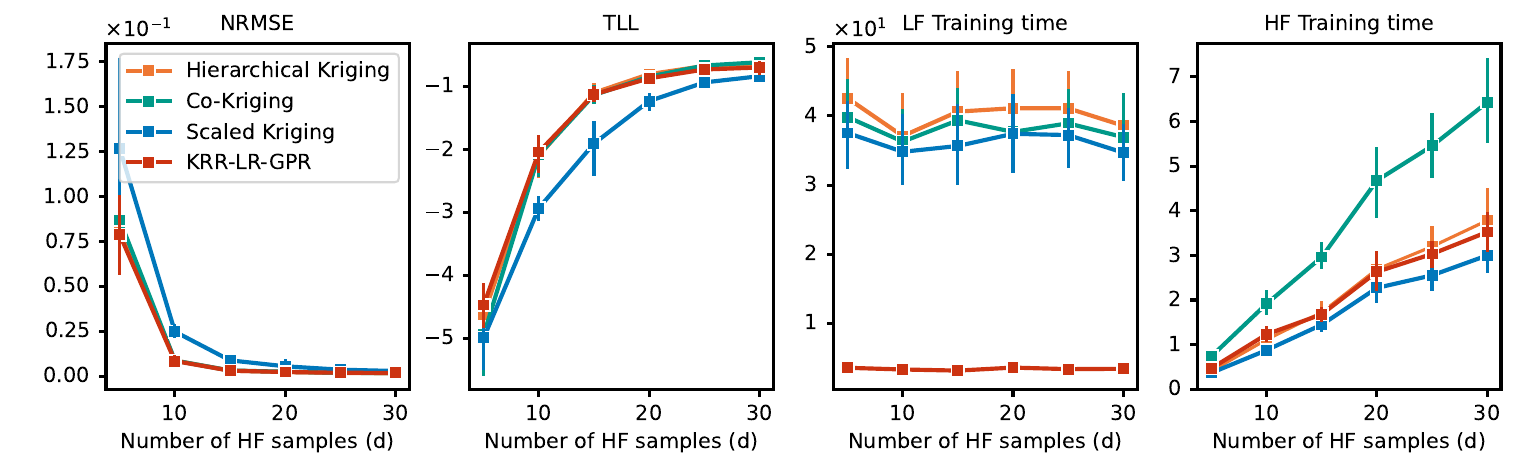}
    \caption{Comparison of KRR-LR-GPR with state-of-the-art MF methods for different sizes of HF datasets when learning the Booth function, and considering $200 \times d$ LF samples (in the case of the Booth function $d=2$, and the Pearson correlation coefficient is $r=0.925$). Different colors represent different methods. To interpret the results, note that smaller NRMSE or higher TLL indicate better performance. Note the significantly faster LF training time for KRR-LR-GPR compared to others.}
    \label{fig: gpr_comparison_at_200lf_samples}
\end{figure}

\begin{figure}[hbt!]
    \centering
    \includegraphics[width=\textwidth]{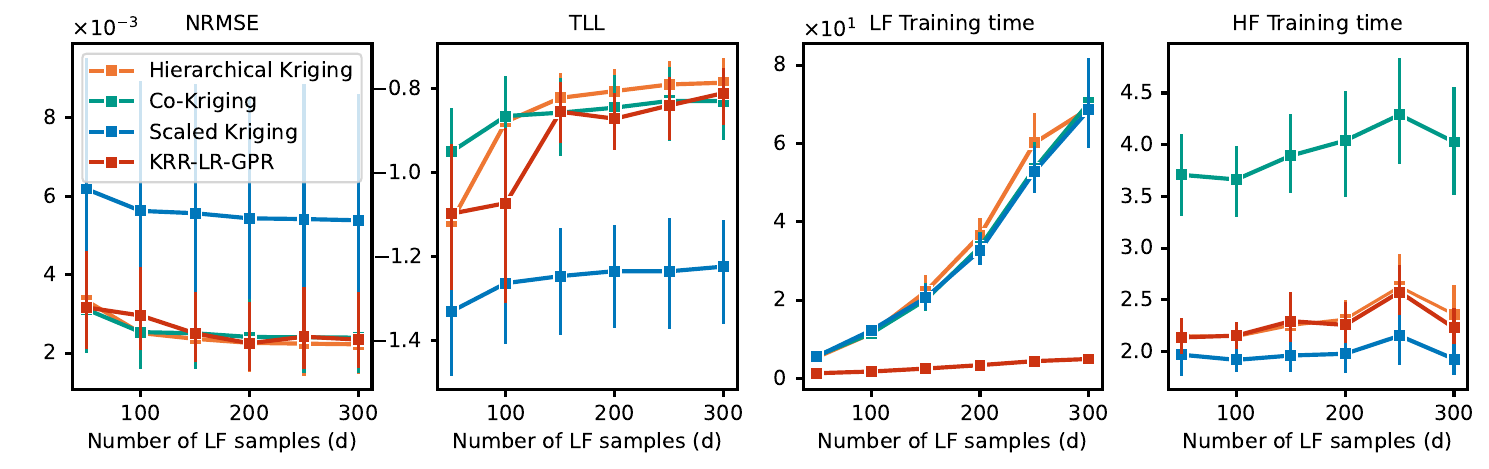}
    \caption{Comparison of KRR-LR-GPR with state-of-the-art methods when learning the \textit{Booth} function but now considering different sizes of the LF dataset while keeping the number of HF samples fixed to $20 \times d$.}
    \label{fig:gpr_comparison_at_20hf_samples} 
\end{figure}

The methods mentioned above are trained on 10 other analytical benchmark functions, reporting results by repeating every experiment with 5 random initializations of the training data generation. All 10 functions lead to similar conclusions; therefore, we arbitrarily select the \textit{Booth} function to show in the main text (results for other functions are reported in \Cref{sec: additional gpr experiments results}). \Cref{fig: gpr_comparison_at_200lf_samples} shows the results assuming a fixed number of LF samples ($200 \times d$ where $d$ is the dimension of the particular function being trained), and then considering a different number of HF samples (from $5 \times d$ to $30 \times d$). \Cref{fig:gpr_comparison_at_20hf_samples} pertains to a different experiment where we fix the number of HF samples to $20 \times d$, and consider different numbers of LF samples (from $50 \times d$ to $300 \times d$). Moreover, Gaussian noise with standard deviations of $\sigma_{l} = 0.3$ and $\sigma_{h} = 0.3$ is considered for the datasets in both cases.

The figures (\Cref{fig: gpr_comparison_at_200lf_samples}, \Cref{fig:gpr_comparison_at_20hf_samples}, \Cref{fig:additional_comprehensive_experiments_200lf}, \Cref{fig:additional_comprehensive_experiments_20hf}) reinforce the argument that considering KRR for the LF and GPR for the residual creates a MF model with comparable performance to considering a GPR for all fidelity levels (hierarchically or not), while the computational cost decreases dramatically. Essentially, the computational complexity is reduced to $\mathcal{O}((N^h)^3 +(N^l)^2 )$ instead of $\mathcal{O}((N^h+N^l)^3)$. For cases where the HF training data is scarce and low dimensional, we recommend using a Bayesian method like GPR for that fidelity and a scalable method like KRR for the LF. It is noted that we also conducted experiments with different HF noise levels in \Cref{sec: ablation study of KRR-LR-GPRR} as well as an ablation study on changing both LF and HF samples in \Cref{sec:additional experiments for KRR-LR-GPR}, those results also support the above arguments. 

\subsubsection{Engineering application: NACA 0012 airfoil}
\label{sec:naca0012_airfoil}

\begin{figure}[h]
    \centering
    \includegraphics[width=\textwidth]{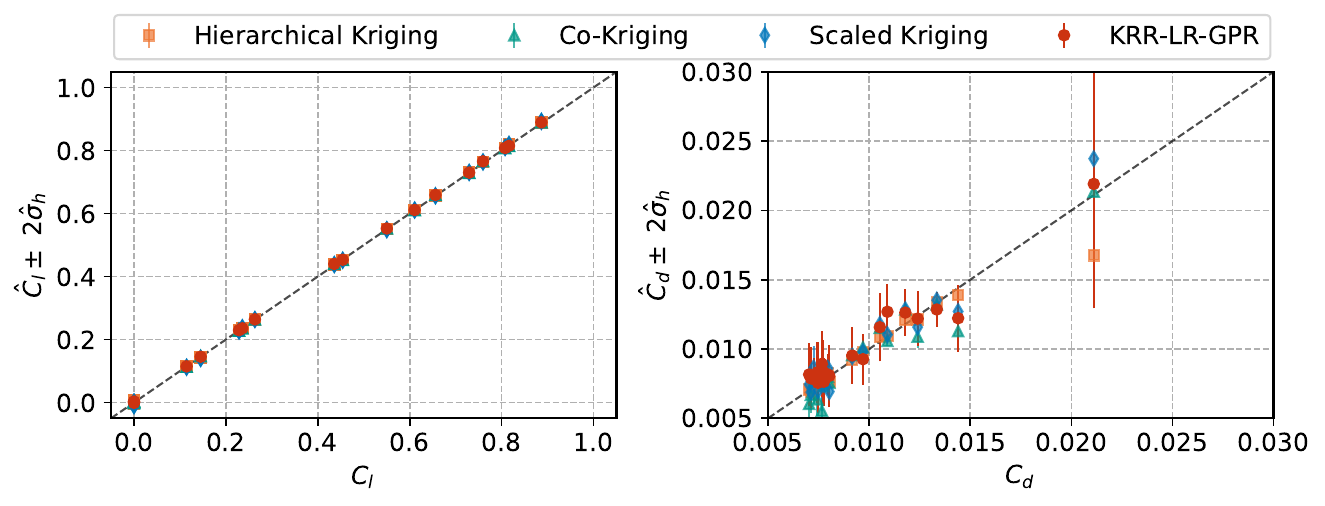}
    \caption{Comparison results of the NACA0012 airfoil. The error bars of each point in the plot indicate predicted uncertainty}
    \label{fig:comparison_results_of_NACA_airfoil}
\end{figure}

The KRR-LR-GPR method is also used to determine the aerodynamic coefficients of a NACA 0012 airfoil, a common engineering problem for assessing MF models \cite{Lin2021, liao2021multi}. The aim is to predict the lift and drag coefficients ($C_l$ and $C_d$) considering two independent variables (freestream Mach number $\text{MA}$ and angle of attack $\text{AoA}$). The dataset was obtained from \cite{Lin2021}, in which 5 HF points (simulated by Reynolds-averaged Navier Stokes equation) and 40 LF points (simulated by Euler equation) are provided for training, and 20 HF points are used for testing. \Cref{tab:results_of_NACA_airfoil} presents the results, and \Cref{fig:comparison_results_of_NACA_airfoil} offers a comparison between all selected data-scarce MF methods.

\begin{table}[h]
    \centering
    \caption{Results of different data-scarce methods for the NACA0012 airfoil MF problem for the two QoIs.}
    \label{tab:results_of_NACA_airfoil}
    \renewcommand{\arraystretch}{1.2}
    \begin{tabular}{c c c c c c c}
        \hline
        \multirow{2}{*}{QoI} &   \multirow{2}{*}{Methods}  &  \multirow{2}{*}{NRMSE}   & \multirow{2}{*}{R2 Score}     &  \multirow{2}{*}{TLL}  & \multicolumn{2}{c}{Training time ($s$)}  
        \\        
         & & & & & HF  & LF \\
        \hline
                             & Hierarchical Kriging  & 0.0058 & \textbf{0.9999} & 0.1636 & 0.0686 & 1.6828  \\
                             & Co-Kriging            & 0.0065 & \textbf{0.9999} & 4.2375 & 0.2138 & 1.4089 \\
        $C_l$                   & Scaled Kriging        & 0.0085 & \textbf{0.9999} & 3.3848 & 0.1270 & 2.1036  \\
                             & KRR-LR-GPR            & \textbf{0.0057} & \textbf{0.9999} & \textbf{4.3414} & \textbf{0.0258} & \textbf{1.0354}\\
        \hline
                             & Hierarchical Kriging  & 0.1038 & 0.9107 & \textbf{5.9189} & 0.1259 & 1.2885 \\
                             & Co-Kriging            & 0.1089 & 0.9016 &4.7347 & 0.1214 & 1.3689  \\
        $C_d$                   & Scaled Kriging        & 0.0987& 0.9192 & 5.2907 & 0.0799 & 1.4836  \\
                             & KRR-LR-GPR            & \textbf{0.0923} & \textbf{0.9293} & 5.5690 &  \textbf{0.0674} & \textbf{0.0745}  \\
        \hline
    \end{tabular}
\end{table}

\Cref{tab:results_of_NACA_airfoil} shows that the performance of the proposed KRR-LR-GPR is better than other methods, especially when predicting $C_d$ but also for $C_l$ (observe the NRMSE values). The TLL value when predicting $C_d$ is the only metric that is marginally worse than Hierarchical Kriging. We note that KRR-LR-GPR is the fastest method to train, although this is not beneficial in this example because the HF and LF data acquisition time is much larger than the time needed to train any of the MF algorithms. In fact, this example is data-scarce even in the LF (only 40 points), and so the better scalability of KRR for LF regression is not needed in this case (this only becomes relevant when the LF dataset has over a few thousand points).

\subsection{Experiments with data-rich HF and high-dimensional problems: DNN-LR-BNN}
\subsubsection{Illustrative example} \label{sec: 1d_mf_dnn_bnn}

\begin{figure} [h]
    \centering
    \includegraphics[width=\textwidth]{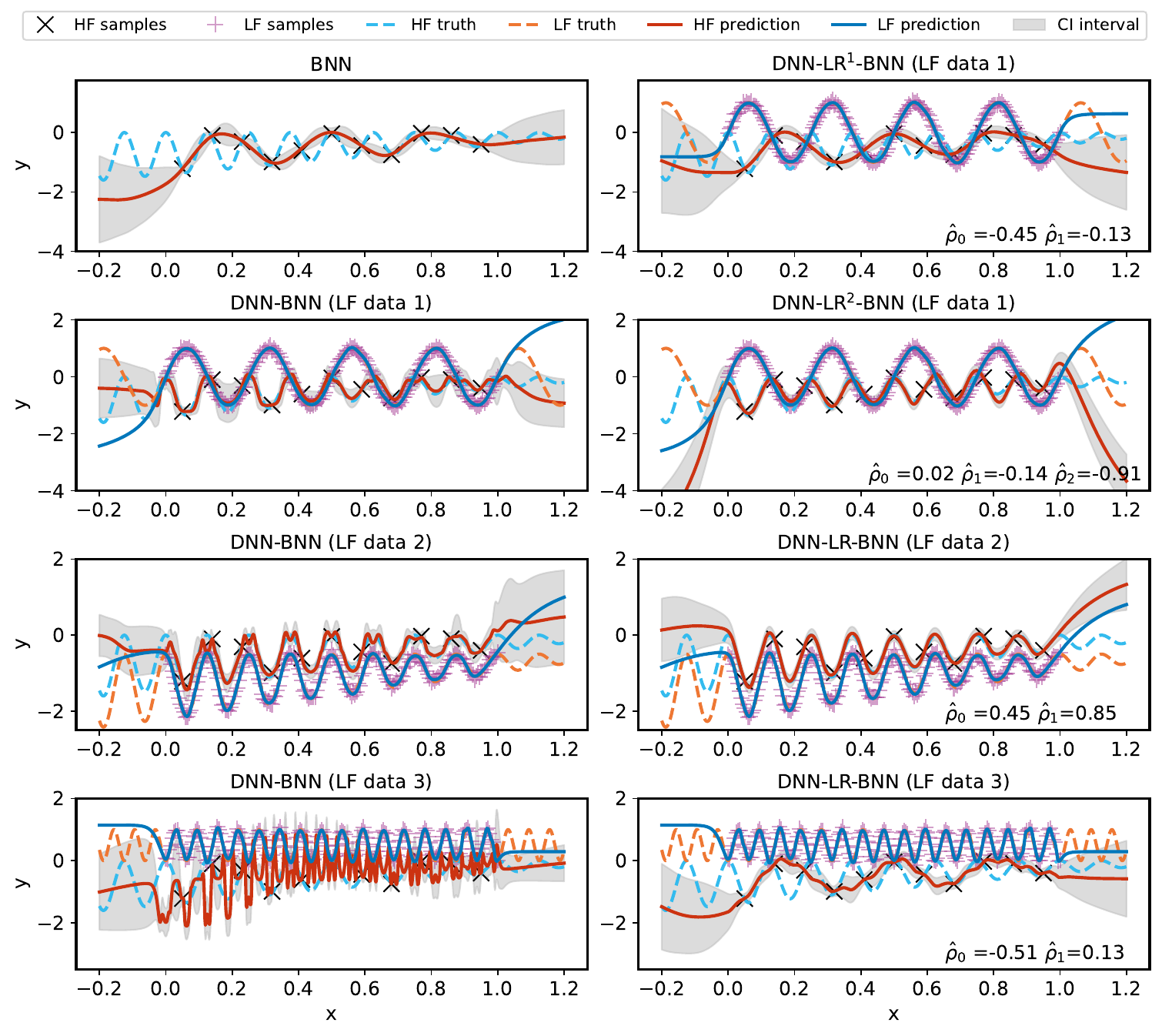}
    \caption{One-dimensional illustrative example comparing the proposed DNN-LR-BNN method (right column) with a BNN trained exclusively with HF data, and with a simple DNN-BNN (left column). Three LF datasets are employed whose Pearson correlation coefficients are $-0.1019$, $0.9999$, and $-0.0018$ correspondingly.}
    \label{fig: mfdnnbnn 1d fitting}
\end{figure}

We start with a one-dimensional illustrative example (data-scarce and low-dimensional) with 11 HF samples and 201 LF samples distributed uniformly. Aiming at explaining the limitations of the different MF models with DNNs and BNNs, we select three different LF datasets, labeled as ``LF data 1'', ``LF data 2'' and ``LF data 3'', obtained by establishing a different correlation with the HF data according to \Cref{eq: meng_1d_lf1}, (\ref{eq: meng_1d_lf2}) and (\ref{eq: meng_1d_lf3}), respectively. Model performance and accuracy metrics are shown in \Cref{fig: mfdnnbnn 1d fitting} and \Cref{tab: results_1D_noisy_mfdnnbnn(known noise)}. 
\begin{center}
\begin{threeparttable}[h]
    \centering
    \caption{Results comparison of DNN-LR-BNN on the 1D noisy illustrative example with 5 independent runs}
    \label{tab: results_1D_noisy_mfdnnbnn(known noise)}
    \begin{tabular}{m{3.5cm} m{3cm} m{2.5cm} m{2.5cm} m{2.5cm}}
        \hline \noalign{\vskip 2mm}
     LF Dataset &     Methods    & NRMSE    & R2 Score & TLL \\ \noalign{\vskip 2mm} \hline \noalign{\vskip 2mm}
       --- &  BNN         & 1.0470 ($\pm 0.028$) & -0.6888 ($\pm 0.093$)  & -5.7825 ($\pm 0.578$)	   \\ \noalign{\vskip 2mm}    
    \multirow{3}{*}{LF Data 1
    ($r$ = -0.1019)}  &   DNN-BNN & 0.4865 ($\pm 0.163$)& 0.6029 ($\pm 0.247$) & \textbf{0.2808} ($\pm0.308$) \\
        & DNN-LR$^1$-BNN & 1.0501 ($\pm 0.013$) & -0.6983 ($\pm 0.043$) & -6.4183 ($\pm 0.662$) \\
            & DNN-LR$^2$-BNN& \textbf{0.3161} ($\pm 0.005$)& \textbf{0.8461} ($\pm 0.005$) & 0.2329($\pm 0.090$)\\  \noalign{\vskip 2mm}
    \multirow{2}{*}{LF Data 2 ($r$ = 0.9999)} &  DNN-BNN & 0.3743 ($\pm 0.092$)& 0.7738 ($\pm 0.102$) & 0.5946 ($\pm 0.136$)\\ 
    & DNN-LR$^1$-BNN & \textbf{0.1051} ($\pm 0.005$) & \textbf{0.9830} ($\pm 0.001$) & \textbf{1.2296} ($\pm 0.018$) \\ \noalign{\vskip 2mm}

    \multirow{2}{*}{LF Data 3 ($r$ = -0.0018)} &   DNN-BNN &1.1012 ($\pm 0.115$) &-0.8834 ($\pm 0.403$)& \textbf{-5.5084} ($\pm 1.899$)\\ 
    & DNN-LR$^1$-BNN & \textbf{1.0643} ($\pm 0.012$)& \textbf{-0.7445} ($\pm 0.038$) & -6.1546 ($\pm 0.711$) \\ \noalign{\vskip 2mm}
     \hline
    \end{tabular}
    \begin{tablenotes}
        \item [*] The subscripts 1 and 2 indicate that the LR model used for transfer-learning is order 1 (linear polynomial basis functions) and order 2 (quadratic polynomial basis functions), respectively.
    \end{tablenotes}
\end{threeparttable}
\end{center}

The datasets \textit{LF data 1} and \textit{LF data 3} were deliberately created with a low Pearson correlation coefficient concerning the HF dataset.  \Cref{fig: mfdnnbnn 1d fitting} and \Cref{tab: results_1D_noisy_mfdnnbnn(known noise)} show that for such cases the LR transfer-learning model used in DNN-LR-BNN automatically finds low values for the $\rho_1$ coefficient after training the LF model on the HF data (independently of the BNN training on the residual). We believe that this is a strength of the proposed strategy because it automatically adjusts the influence of the LF model and informs the analyst about the correlation between LF and HF data. We note that the order of the LR transfer-learning model is a hyperparameter, but the coefficients $\boldsymbol{\rho}$ are found by training via minimizing \Cref{eq: obj of hyperparameter}. Simple hyperparameter search strategies can quickly find the optimal order of the LR transfer-learning model for each problem (it is only one hyperparameter). For example, a quadratic LR transfer-learning model is better than a linear one for ``LF data 1'' (this is labeled as DNN-LR$^2$-BNN to distinguish from the LR model with linear basis functions labeled as DNN-LR$^1$-BNN). Evidently, if LF and HF data tend to be linearly correlated (as in the \textit{LF data 2}), the DNN-LR-BNN model performs better.

Interestingly, even when the correlation between LF data and HF data is highly nonlinear, we have not found the performance of the DNN-BNN to be better, despite this model using a BNN to perform transfer-learning instead of a simple LR model. The next sections demonstrate this for different datasets, including a material modeling example that illustrates a more realistic scenario encountered in engineering practice. This might explain why early MF literature considered first-order linear relationships between LF and HF models that were governed by a single hyperparameter ($\rho_1$). In essence, we find that performing model selection and establishing a simple (but not simpler) transfer-learning model between the DNN and BNN is an important consideration.

We also want to bring to the reader's attention that neural network training is highly dependent on architectural and hyperparametric choices. \Cref{sec:hyperparameter-study} details the dependence on hyperparameters and reinforces that the DNN-LR-BNN model is simpler to train and is less sensitive to hyperparameter choices than the more flexible DNN-BNN model. We observed in all cases that the prediction of uncertainty depends on the width of the BNN, i.e. on the number of neurons in the hidden layers. A wide network leads to similar uncertainty predictions as the ones obtained by GPR, while the prediction of the expected value (the mean) converges quickly even for narrow BNNs.

\subsubsection{Higher dimensional numerical examples}
\label{sec:dnn_lr_bnn_higher_dimensional_examples}

We also demonstrate the performance of DNN-LR-BNN and DNN-BNN in problems with higher dimensions and larger datasets. We consider two functions, a 4-dimensional function extracted from \cite{Meng2021_MFBNN} and another function from \cite{Meng2020_MFPINN} where we set the dimension to be [20, 50, 100] in this subsection. The results are summarized in  \Cref{tab:results_noisy_mfdnnbnn}. 

\begin{table}[h]
    \centering
    \caption{Results comparison of DNN-LR-BNN on the high-dimensional examples}
    \label{tab:results_noisy_mfdnnbnn}
    \begin{tabular}{m{1.5cm} m{2.5cm} m{1.5cm} m{1.5cm} m{2cm} m{1.5cm} m{1.5cm}}
        \hline
     Dimension  &   Methods       & NRMSE    & R2 Score & TLL & $\hat{\rho}_0$& $\hat{\rho}_1$ \\
        \hline
         
     \multirow{3}{*}{4} &     BNN       & 0.7624 ($\pm 0.031$) & -0.6296 (0.131)  & -5.3555 ($\pm 0.458$)  & - & -   \\
   &     DNN-BNN & 0.1163 ($\pm 0.013$) & 0.9617 ($\pm 0.009$) & 1.2680 ($\pm 0.040$) & - & - \\
     &   DNN-LR$^1$-BNN  & \textbf{0.0777} ($\pm 0.013$)  & \textbf{0.9826} ($\pm 0.006$) & \textbf{1.4090} ($\pm 0.045$)  & -0.41 ($\pm 0.002$) &0.86 ($\pm 0.006$)  \\ 
        \hline
      \multirow{3}{*}{20} &   BNN        & 0.1598 ($\pm 0.004$) & 0.7418  ($ \pm 0.013 $ )  &-23.5060 ($\pm 1.673$)  & - & -   \\
       &DNN-BNN  & 0.0723 ($\pm 0.010$) & 0.9463 ($\pm 0.015$) & \textbf{-8.1125} ($\pm 1.137$) & - & - \\
        & DNN-LR$^1$-BNN  & \textbf{0.0662} ($\pm 0.008$)  & \textbf{0.9550} ($\pm 0.012$) & -8.7991 ($\pm 0.905$)  & 55.75 ($\pm 17.091$) & 1.26 ($\pm 0.005$)  \\ 
        \hline
              \multirow{3}{*}{50} &   BNN        & 0.1725 ($\pm 0.003$) & 0.2341  ($ \pm 0.024 $ )  &-59.2736 ($\pm 3.671$)  & - & -   \\
       &DNN-BNN  & 0.0879 ($\pm 0.001$) & 0.8008 ($\pm 0.007$) & -17.7602 ($\pm 0.9644$) & - & - \\
        & DNN-LR$^1$-BNN  & \textbf{0.0851} ($\pm 0.001$)  & \textbf{0.8137} ($\pm 0.003$) & \textbf{-17.3543} ($\pm 0.0.8955$)  & 62.0892 ($\pm 5.219$) & 1.2504 ($\pm 0.001$)  \\ 
        \hline
              \multirow{3}{*}{100} &   BNN        & 0.1343 ($\pm 0.001$) & 0.0564  ($ \pm 0.020 $ )  &-113.4960 ($\pm 8.007$)  & - & -   \\
       &DNN-BNN  & 0.0867 ($\pm 0.002$) & 0.6062 ($\pm 0.018$) & -58.8356 ($\pm 6.7228$) & - & - \\
        & DNN-LR$^1$-BNN  & \textbf{0.0845} ($\pm 0.001$)  & \textbf{0.6265} ($\pm 0.008$) & \textbf{-58.7305} ($\pm 4.458$)  & -100.0 ($\pm 0.0$) & 1.2868 ($\pm 0.013$)  \\ 
        \hline

    \end{tabular}
\end{table}

\Cref{tab:results_noisy_mfdnnbnn} shows that the single-fidelity BNN  leads to poor predictions for all examples according to all accuracy metrics. Also for all cases, the DNN-LR-BNN outperforms the DNN-BNN model. The coefficients $\hat{\bm{\rho}}$ learned for the DNN-LR-BNN model are reasonable when compared to the ground-truth functions shown in \Cref{sec: high dimensional functions}.

\subsubsection{Engineering application: material structure-property prediction dataset}
\label{sec:mat_law_prediction}
 
\begin{figure}
    \centering
    \includegraphics[width=\textwidth]{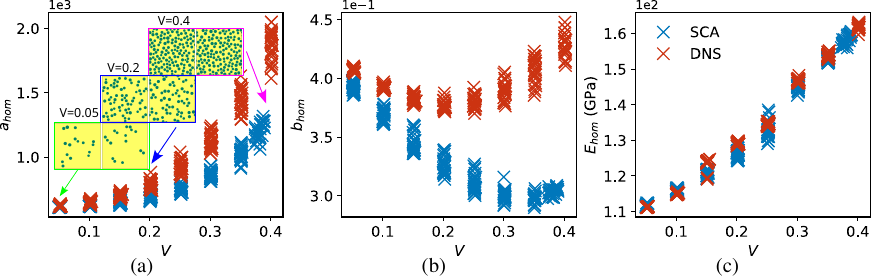}
    \caption{Material structure-property prediction problem illustration. This problem involves four input descriptors: volume fraction $V$ of the particle material, Young's modulus of the linear elastic fibers $E_\texttt{fiber}$, matrix hardening law coefficients $\sigma = 400 + a_\text{matrix}\varepsilon_p^{b_\text{matrix}}$. The simulations of the material response lead to three outputs: homogenized Young's modulus $E_\text{hom}$, and homogenized hardening law with coefficients $a_\text{hom}$  and $b_\text{hom}$. We utilize two fidelity levels for data generation: HF data is generated through a FEA method via commercial software (ABAQUS) based on \cite{yi2023rvesimulator} and the LF data is obtained via the SCA method from the CRATE open source package \cite{ferreira2023crate}. Figures (a), (b), and (c) show three examples of how one output property changes concerning a particular input variable, making it clear that there is aleatoric uncertainty (noise).  \Cref{sec:hyper_parms_mat_law} contains additional information about the data generation process for this problem.}
    \label{fig:mat_structrue_property_illustration}
\end{figure}

\begin{figure}[h]
    \centering
    \includegraphics[width=\textwidth]{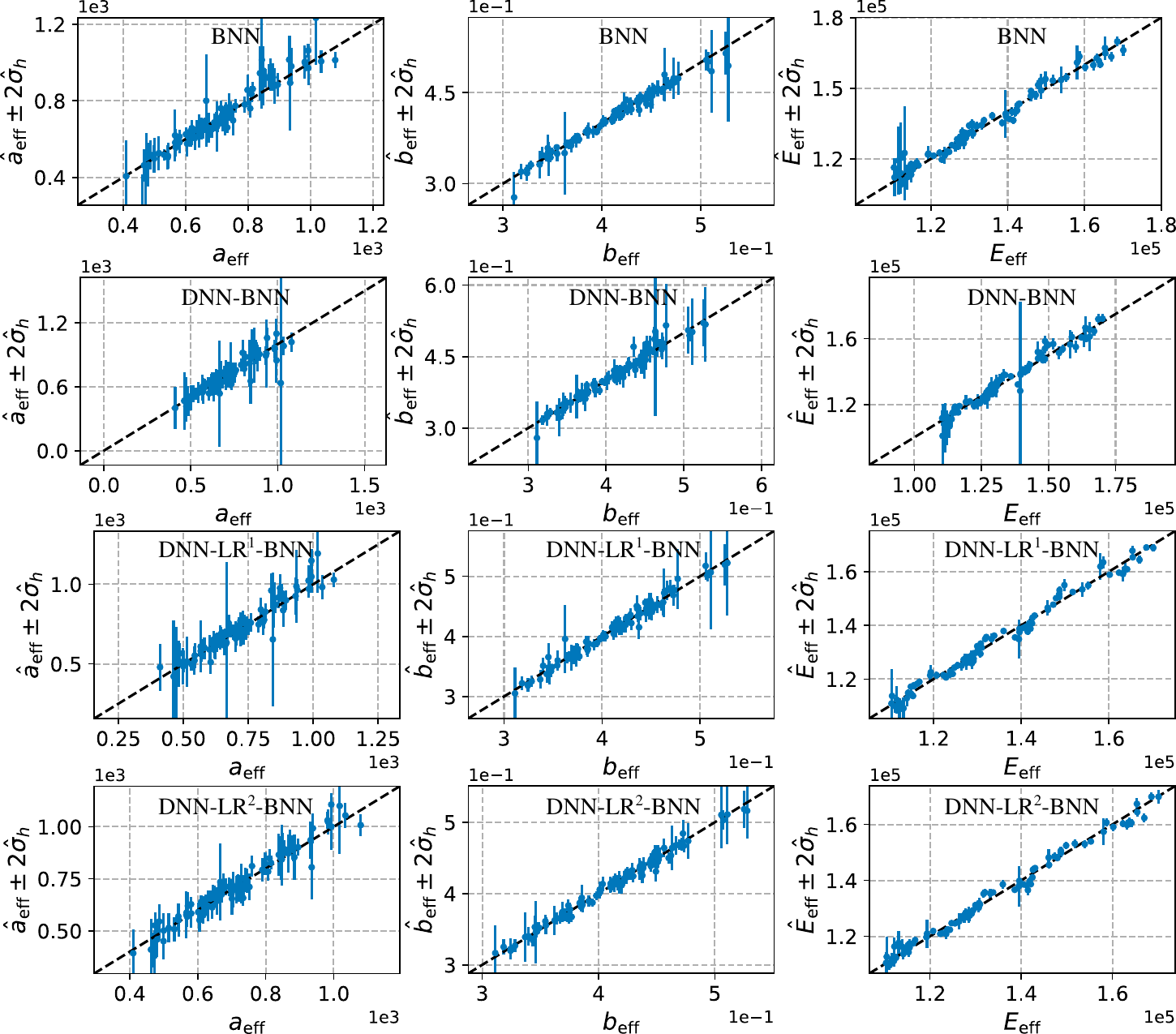}
    \caption{Predicted outputs with uncertainties for material structure-property linkages problem. The size of the points represents the predicted uncertainty }
    \label{fig:mat_structure_property_linkage}
\end{figure}

This section focuses on finding structure-property relationships in materials, as described in \cite{bessa2017framework}. The dataset considered here was introduced by Olivier et al. \cite{Olivier2021}, where effective material properties of two-phase material microstructures are determined by computer simulations of material samples, called SVEs. Each SVE results from randomizing the material microstructure, then deforming the SVE according to periodic boundary conditions, and susbequently calculate the average (or homogenized) response of the SVE. The authors of that study only considered outputs from a single-fidelity, obtained by direct numerical analysis (DNS) via finite element analysis (FEA), and showed that a BNN can predict both the mean and standard deviation for chosen quantities of interest (QoIs). We extend this problem to the MF scenario by generating LF data using an open-source implementation \cite{ferreira2023crate} of a significantly faster simulation method called self-consistent clustering analysis (SCA) \cite{Liu2016,ferreira2022adaptivity}.  \Cref{fig:mat_structrue_property_illustration} summarizes the material structure-property regression problem, and additional information can be found in the original publication \cite{Olivier2021}. We highlight that a reader lacking domain knowledge of the Mechanics of Materials and multi-scale simulations may choose to ignore the physical meaning of the input and output variables of this dataset. In that case, the reader only needs to know that there are four input variables with the following bounds:
\begin{align*}
    V &\in [0.05, 0.40], \\
    E_{\text{fiber}} &\in [200000, 600000], \\
    a_{\text{matrix}} &\in [300,500], \\
    b_{\text{matrix}} &\in [0.2, 0.55],
\end{align*}
\noindent that lead to noisy measurements captured by three output variables: $E_\text{hom}$, $a_\text{hom}$ and $b_\text{hom}$, where the subscript ``hom'' indicates that these properties correspond to the homogenized (average) response of the two-phase material. The LF output measurements are obtained by a fast but less accurate method, while the HF measurements of the same three output variables are obtained by a slower but more accurate method. Note that both LF and HF measurements are noisy, as the same input point leads to different output values each time it is evaluated (due to material microstructure randomness). \Cref{fig:mat_structure_property_linkage} illustrates the variation of one of the outputs with respect to one of the inputs, while keeping the remaining inputs fixed. In that figure, red crosses correspond to HF data (DNS) and blue crosses to LF data (SCA). 

We generated 500 HF samples with the FEA method and split them into 400 for training and 100 for testing. In addition to this, we also generate 3200 LF samples using the SCA method (using 3 material clusters -- the hyperparameter of SCA that controls discretization). We then test the proposed DNN-LR-BNN strategy on this problem where we use 300 HF samples and 3200 LF samples to train the MF models, i.e. DNN-BNN and DNN-LR-BNN. We also train a single-fidelity BNN model using 400 HF samples, as this corresponds to the number of samples that can be generated with the same computational resources and time needed to create the MF dataset (in other words, 100 HF simulations generated with FEA require approximately the same time as 3200 LF simulations with the SCA method).  \Cref{tab:mat_structure_property_linkage} compares the performance of the different MF models, including the single-fidelity baseline of BNN.  \Cref{fig:mat_structure_property_linkage} presents the output predictions compared to the measured output values in the test set. However, recall that the outputs are noisy and that their ground-truth mean values are not known, nor their corresponding aleatoric distribution (noise distribution). Therefore, the figure only shows the mean prediction (y-axis value), and corresponding epistemic uncertainty estimation (vertical bar) for each test sample (x-axis value). 

\begin{table}[h]
    \centering
    \caption{Results comparison of DNN-LR-BNN on the material structure-property linkages}
    \label{tab:mat_structure_property_linkage}
    \renewcommand{\arraystretch}{1.2}
    \begin{tabular}{m{2.5cm} m{2.5cm} m{1.5cm} m{1.5cm} m{1.5cm} m{1cm} m{0.6cm} m{2cm}}
        \hline
     QoIs  &   Methods       & NRMSE    & R2 Score & TLL & $\hat{\rho}_0$& $\hat{\rho}_1$ & $\hat{\rho}_2$ \\
        \hline
         
     \multirow{4}{*}{$a_\text{eff}$ ($r$ = 0.9294)} &     BNN       & 0.0902  & 0.8165  & -5.3577   & - & -   \\
      &     DNN-BNN  & 0.1254  & 0.6450 & -5.2192  & - & - \\
     &   DNN-LR$^1$-BNN  & 0.0700 & 0.8894 & -5.3748  & 10.00 & 0.92 & - \\ 
          &   DNN-LR$^2$-BNN  & \textbf{0.0442} & \textbf{0.9558} & \textbf{-4.9248}  & 10.00 & 1.24 & -3.55 $\times 10^{-4}$ \\ 
        \hline
      \multirow{4}{*}{$b_\text{eff}$ ($r$ = 0.7681)} &   BNN        & 0.0268 & 0.9489   &3.2622  & - & -   \\
       &DNN-BNN   & 0.0345 & 0.9155  &3.1881 & - & - \\
        & DNN-LR$^1$-BNN  & 0.0178 & 0.9774 & 3.3436  & 0.10 & 0.88 & -  \\ 
        & DNN-LR$^2$-BNN  & \textbf{0.0134} & \textbf{0.9872} & \textbf{3.5999} & 0.05 & 1.11 & 0.32  \\ 
        \hline
              \multirow{4}{*}{$E_\text{eff}$ ($r$ = 0.9845)} &   BNN        & 0.0025  & 0.9556   & -10.8660   & - & -   \\
              &DNN-BNN & 0.0229& 0.9649 & -10.2588& - & - \\
        & DNN-LR$^1$-BNN  & 0.0146  & 0.9857 & -9.2766 & -7.97& 1.01 & -   \\
        & DNN-LR$^2$-BNN  & \textbf{0.0121}  & \textbf{0.9901} & \textbf{-9.2447}  & -1.27& 0.99 & -7.75 $\times 10^{-8}$  \\
        \hline

    \end{tabular}
\end{table}

\Cref{tab:mat_structure_property_linkage} compares different MF models and also includes the single-fidelity BNN prediction. This table clarifies that the proposed DNN-LR-BNN model with a quadratic linear regression transfer learning model (DNN-LR$^2$-BNN) has the best performance for this material structure-property problem.  These results demonstrate the effectiveness of the DNN-LR-BNN method in real-world scenarios where the dataset contains multiple outputs.  \Cref{fig:mat_structure_property_linkage} also highlights that the single-fidelity BNN has the largest predicted uncertainty for all test points. Conversely, the DNN-LR-BNN has a more reasonable uncertainty prediction, achieving the best TLL values among all methods. We can see that in cases where the outputs of the two fidelity levels have a high Pearson correlation coefficient ($r \rightarrow 1$), then the performance is similar for DNN-LR$^1$-BNN and DNN-LR$^2$-BNN because the training procedure identifies that the quadratic coefficient of the transfer-learning model in DNN-LR$^2$-BNN tends to zero ($\hat{\rho}_2 \rightarrow 0$). Instead, for the $b_\text{hom}$ output where the two fidelity levels are not as linearly correlated, the quadratic transfer learning model improves the prediction, as it facilitates training the BNN on the HF data (note that in this case we have $\hat{\rho}_2 \neq 0$). We experienced difficulties in avoiding overfitting when training the DNN-BNN method for some of the outputs in this problem.

\section{Discussion and limitations} \label{sec: empirical proof}

\begin{figure}[h]
    \centering
    \includegraphics[width=\textwidth]{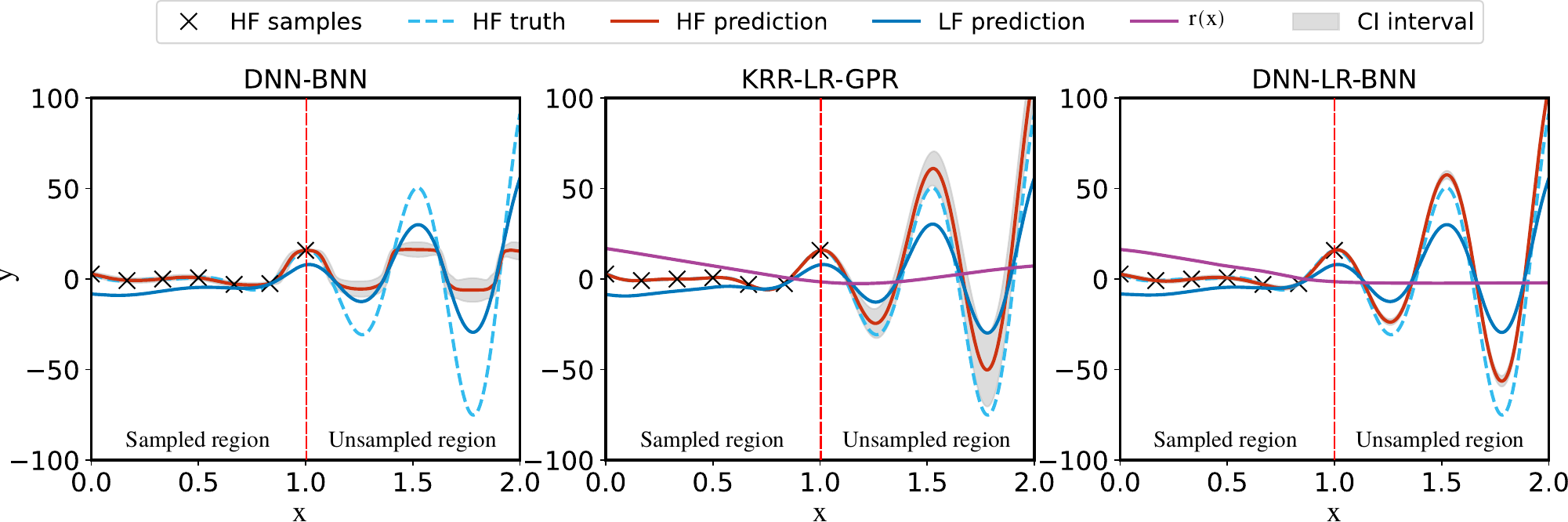}
    \caption{Illustration of what occurs to the DNN-BNN, KRR-LR-GPR and DNN-LR-BNN model predictions when there is a domain region without HF samples ($x>1.0$), but considering LF samples throughout the entire domain ($0<x<2$).}
    \label{fig: empirical proof on equivalence of MF-BML}
\end{figure}

Expressive machine learning algorithms typically have limited extrapolation ability. Their large number of parameters is used to explain the training data without necessarily discovering the underlying functional form that explains the data. DNNs often make unreliable predictions for points away from the support domain; conversely, Bayesian models such as GPRs and BNNs eventually reverse to the prior \cite{Adriaensen2023}. A similar issue also occurs in the DNN-BNN architecture where the HF prediction is distorted in the unsampled region as reported in  \Cref{fig: empirical proof on equivalence of MF-BML}. However, KRR-LR-GPR and DNN-LR-BNN do not exhibit this behavior because they remain smooth and close to the HF truth. The proposed transfer-learning model in \Cref{eq: mf-bml} narrows the gap between $\mathbf{m}(\mathbf{X}^h)^T\bm{\rho}$ and $\mathbf{y}^h$, while the BNN captures the residual utilizing a prior with zero mean that reverses to another constant after training (see purple line in \Cref{fig: empirical proof on equivalence of MF-BML}). Therefore, the HF prediction is influenced by the parsimonious (simple) transfer-learning model, improving its extrapolation ability.

However, when we compare the DNN-LR-BNN model prediction with the KRR-LR-GPR one, we see that although the mean is similar, the uncertainty prediction is not. The predicted uncertainty of DNN-LR-BNN is smaller in the extrapolation region -- the model is overconfident. We identify two reasons for this observation. First, BNN inference is approximate and does not yield the exact posterior predictive distribution. This bottleneck can be resolved for a specific problem by fine-tuning the BNN hyperparameters using test data or cross-validation, as well as by increasing the number of neurons in the hidden layers. Second, there is an inherent bottleneck in evaluating uncertainty via  \Cref{eq:predicted variance of explicit mean function} when a BNN is employed for $f^h(\mathbf{x})$ inference: the covariance matrix $\mathbf{K(X, X)}$ is difficult to compute and so the uncertainty present in the LF model is difficult to transport to the HF prediction.

\section{Conclusions} \label{sec: conclusions}

In this study, we created a general MF framework that encompasses different MF strategies in the literature. We propose using non-probabilistic LF models together with Bayesian HF models and recommend two practical strategies to deal with data-scarce and data-rich scenarios. For the first case, we propose the KRR-LR-GPR MF model that considers KRR for training on LF data, linear regression for LF to HF data transfer-learning, and Gaussian process regression for modeling the residual that remains. For data-rich, high-dimensional cases and/or multiple outputs, we recommend the DNN-LR-BNN model that involves training a deep neural network on LF data, transfer-learning via linear regression, and Bayesian neural networks for learning the residual. If a linear regression model does not provide an effective transfer-learning method, a more flexible DNN-BNN model is recommended using pSGLD for Bayesian inference.

The KRR-LR-GPR model was assessed for cases with and without noise for both fidelity levels, and its performance was found to be similar to state-of-the-art MF-GPR methods. However, KRR-LR-GPR exhibits two key advantages: 1) it can be trained on larger LF datasets, and 2) it is significantly faster to train (in practice, the method is only bounded by the number of HF samples, due to the cubic complexity of GPR models with the number of training samples). The method was also successfully applied to a dataset from an engineering application -- prediction of aerodynamic coefficients of NACA 0012 airfoil. We believe that the simplicity of the training strategy and robustness of the method makes it suitable for most engineering applications because the most common scenario involves data-scarce HF samples (usually originating from expensive or time-consuming experiments or simulations).

We also proposed the DNN-LR-BNN model and compared it with both a single-fidelity BNN and a flexible MF model composed of a DNN and a BNN (DNN-BNN model). We observed that the DNN-LR-BNN model has a lower tendency to overfit when compared to DNN-BNN because its linear regression (LR) transfer-learning model isolates the HF from the LF when they are not linearly correlated. Furthermore, we highlight that the proposed strategy determines the coefficients of the LR model, $\bm{\rho}$, directly from training (instead of considering them as hyperparameters, as done in the literature for other models). The learned $\hat{\bm{\rho}}$ also indicates the correlation between HF and LF data, as shown in different examples such as the material structure-property relationship problem. Nevertheless, the DNN-LR-BNN model has more hyperparameters than the KRR-LR-GPR model, so it should be selected only when training the latter is not viable.

\section*{Data availability}
The code is available on: \href{https://github.com/bessagroup/mfbml}{https://github.com/bessagroup/mfbml}

\section*{Declaration of Generative AI and AI-assisted technologies in the
writing process}

The first author used CHATGPT and GRAMMARLY to IMPROVE LANGUAGE AND READABILITY. After using these tools/services, the authors reviewed and edited the content as needed and assumed full responsibility for the content of the publication.

\section*{Acknowledgments}
Jiaxiang Yi would like to acknowledge the generous support of the China Scholarship Council (CSC). We also appreciate Dr. Quan Lin for his kindness in providing the source data for the NACA 0012 airfoil problem.

\bibliographystyle{unsrt}  
\bibliography{references} 

\newpage
\appendix

\section{Foundations of Bayesian machine learning} \label{sec: foundations}

\subsection{Bayesian neural networks}
A regression problem can be defined by
\begin{equation}
    \mathbf{y}= f(\mathbf{x}) + \epsilon
\end{equation}
where $\epsilon$ is the data noise originating from the collecting process. In this paper, the data noise is assumed to be homoscedastic and obeys Normal distribution $\epsilon \sim \mathcal{N}(0, \sigma^2)$. 

\begin{figure}[h]
    \centering
    \includegraphics[width=0.60\textwidth]{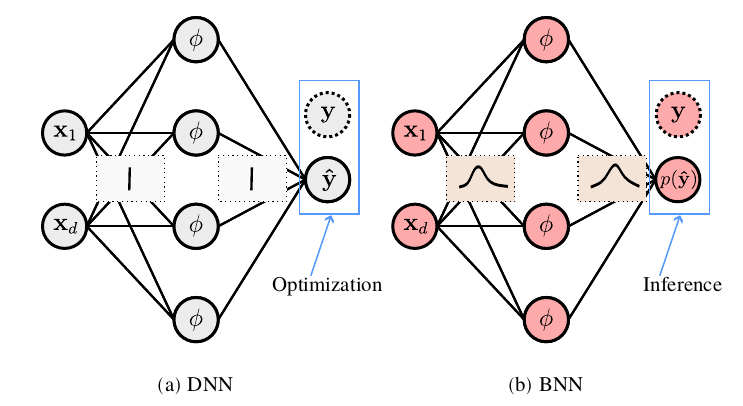}
    \caption{Schematics of DNN and BNN}
    \label{fig:dnn and bnn schematics}
\end{figure}

Consider first the nonprobabilistic DNNs. \Cref{fig:dnn and bnn schematics}.(a) shows a DNN schematic, where the input $\mathbf{x}$ passes through successive neurons located in hidden layers where nonlinear transformations $\phi(\mathbf{x})$ happen in so-called activation functions. To the end, the prediction $\hat{\mathbf{y}}$ is outputted by the DNN and compared to the true data via the loss function. The solid lines that connect neurons include parameters called weights $\mathbf{w}$ while biases are not explicitly included in the schematic (together the weights and biases parameters are labeled as $\boldsymbol{\theta}$). These parameters are real values that are determined via optimization algorithms like Adam \cite{Kingma2014} and Stochastic Gradient Decent \cite{ruder2017overview} that minimize the loss function. 

BNNs are similar to DNNs, as shown in \Cref{fig:dnn and bnn schematics}.(b), but have a key difference: distributions acting on weights are considered instead of using a deterministic value as in DNNs. This probabilistic treatment allows BNNs to capture uncertainty in the predictions. Consequently, traditional optimization algorithms, suitable for finding point estimates, are not directly applicable to BNNs. The procedure to find the posterior parameter distribution is called inference, such that the posterior is determined as follows \cite{Neal1995}: 
\begin{equation} \label{eq: posterior_of_bnn_parameters}
    p \left ( \boldsymbol{\theta} \mid \mathcal{D} \right ) = 
    \frac{  p \left ( \mathcal{D} \mid \boldsymbol{\theta} \right) p \left( \boldsymbol{\theta} \right) }
    { p \left ( \mathcal{D} \right )} , \,\, p \left ( \mathcal{D} \right ) = \int p \left ( \mathcal{D}, \boldsymbol{\theta} \right ) p \left ( \boldsymbol{\theta} \right ) \mathrm{d} \boldsymbol{\theta}
\end{equation}

\noindent where $p \left( \boldsymbol{\theta} \right)$ is the prior, for example, a normal distribution with $\boldsymbol{\theta}$ containing the mean $\boldsymbol{\mu_{\mathbf{w}}}$ and standard deviation of the weights $\boldsymbol{\sigma}_{\mathbf{w}}$;  $p \left ( \mathcal{D} \mid \boldsymbol{\theta} \right)$ is the likelihood; $ p \left ( \mathcal{D} \right )$ is the marginal likelihood or evidence, which is the probability integral of likelihood over $\boldsymbol{\theta}$;  and $ p \left ( \boldsymbol{\theta} \mid \mathcal{D} \right )$ is the posterior distribution of parameter $\boldsymbol{\theta}$.  The aim when training BNNs is to maximize $p \left ( \boldsymbol{\theta} \mid \mathcal{D} \right )$, although this is usually intractable because of the unbounded and complex probability integral. Therefore, methods such as Markov Chain Monte Carlo \cite{Neal1995} and Variational Inference \cite{Blundell2015} are extensively used to determine an approximate posterior distribution. After obtaining parameters, the prediction for an unknown point is calculated by: 
\begin{equation} \label{eq: ppd_of_bnn}
        \hat{p} \left (\mathbf{y}^{\prime}\mid \mathbf{x^\prime}, \mathcal{D} \right )
    = \int p \left ( \mathbf{y^\prime} \mid \mathbf{x^\prime}, \boldsymbol{\theta}\right)  p \left ( \boldsymbol{\theta} \mid \mathcal{D} \right ) \mathrm{d} \boldsymbol{\theta}
\end{equation}
where $\mathbf{x^\prime}$ is the location of the unknown point.

\subsubsection{Bayesian inference method: pSGLD} \label{sec:pSGLD} 

As the analytical solution of the posterior predicted distribution of BNNs is intractable, we used pSGLD \cite{Li2015}. The posterior of a BNN can be regarded as $p(\bm{\theta} | \mathcal{D}) \propto p(\bm{\theta})\prod_{n=1}^{N} p(y_i | \bm{\theta}) $ according to \Cref{eq: posterior_of_bnn_parameters}, and we can define the update rule for the posterior as in SGLD \cite{Welling2011}:
\begin{equation} \label{eq: sgld}
    \Delta \theta_t = \frac{\bm{\epsilon}_t}{2} \left(\nabla \log p(\bm{\theta}_t )  +\frac{N}{n} \log p(y_i|\bm{\theta}_t)\right) + \bm{\eta}_t 
 \end{equation}
where  $\bm{\eta}_t \sim \mathcal{N}(\bm{0}, \bm{\epsilon}_t)$.  However, SGLD assumes that all parameters $\bm{\theta}$ have the same step size, which leads to slow convergence or even divergence in the cases where the components of $\bm{\theta}$ have different curvature. In pSGLD, the update rule incorporates a user-defined preconditioning matrix $G(\bm{\theta}_t)$, which adjusts the gradient updates and the noise term adaptively:

\begin{equation} \label{eq: sgld}
    \Delta \theta_t = \frac{\bm{\epsilon}_t}{2} \left[G(\bm{\theta}_t) \left(\nabla \log p(\bm{\theta}_t )  +\frac{N}{n} \log p(y_i|\bm{\theta}_t)\right)  + \Gamma(\bm{\theta}_t) \right ] + \bm{\eta}_tG(\bm{\theta}_t)
 \end{equation}
where  $\Gamma_i = \sum_j \frac{\partial G_{i,j}(\bm{\theta}) }{\partial \theta_j}$ describes how the preconditioner changes with respect to $\bm{\theta}$. 

We can then obtain the posterior approximately with a set of $\bm{\Theta} = [\bm{\theta}_1,\bm{\theta}_2, ..., \bm{\theta}_n ] $ after $n$ updates from pSGLD. Then, the approximation of \Cref{eq: ppd_of_bnn} can be obtained by a Bayesian model average \cite{Jospin2022}.

\subsection{Gaussian process regression} \label{sec: gpr introduction}

\subsubsection{Gaussian process regression with explicit basis function} \label{sec: explicit gp}
GPR is a kernel machine learning method that can be viewed as the generalized form of Bayesian linear regression or a BNN with one infinite hidden layer, where the prior distribution is a standard Normal distribution \cite{Rasmussen2006}. A general format of a GPR can be expressed as follows when using the function space view and assuming a mean function given by a linear model (on the weights $\bm{\beta}$):
\begin{equation} \label{eq: gp with explicit mean}
f(\mathbf{x}) =
    \mathbf{h}(\mathbf{x})^{T}\boldsymbol{\beta} + \delta(\mathbf{x})
\end{equation}
where $\mathbf{h}(\mathbf{x})$ is a set of basis function, i.e. $\mathbf{h}(\mathbf{x}) = [1, \mathbf{x}, \mathbf{x}^2, ...]^T$ where the length is set to be $M$; $\delta(\mathbf{x})$ is a GPR with zero mean expressed as $\mathcal{GP}(0, k(\mathbf{x}, \mathbf{x}))$, $\bm{\beta}$ is a set of additional parameters that used to tune basis function for minimizing the residual between $\mathbf{h}(\mathbf{x})^{T}\bm{\beta}$ and $\mathbf{y}$ (see \Cref{fig:mf_bml_framework}). The parameters within the covariance function and $\bm{\beta}$ are determined by
maximum likelihood estimation referred to  \Cref{sec: params estimation of GPR}. The predictions\footnote{For
simplification, \Cref{eq:predicted mean of explicit mean function} and
\ref{eq:predicted variance of explicit mean function} are derived assuming
that $\bm{\beta}$ is a deterministic value rather than another distribution. Readers can find more information in \cite{Rasmussen2006} for cases where
$\bm{\beta}$ also follows a distribution.} of the general GPR considering the effect brought by the explicit basis function can then be expressed as\footnote{ \Cref{eq:predicted mean of explicit mean function} and
\ref{eq:predicted variance of explicit mean function} assume no noise in observations. If the response has noise that is independently identically distributed and Gaussian $\varepsilon$ with variance $\sigma^2$, the kernel function should have one extra term to characterize noise. To this end, the noise kernel function can be written as $\mathbf{K}_{noisy}\left( \mathbf{X},
        \mathbf{X}\right ) = \mathbf{K}\left( \mathbf{X}, \mathbf{X}\right ) +
        \sigma^2\mathbf{I}$} \footnote{Naming convention: $k(\mathbf{x}, \mathbf{x})$ represents a number; $\mathbf{k}(\mathbf{x}, \mathbf{X})$ represents a $1 \times N$ vector and vice verse;  $\mathbf{K}(\mathbf{X}, \mathbf{X})$ is a $N \times N$ matrix. Moreover, the default vector is a column vector.}:. 
\begin{equation} \label{eq:predicted mean of explicit mean function}
    \hat{f}(\mathbf{x}^{\prime})=  \mathbf{h}(\mathbf{x^\prime})^{T}\mathbf{\hat{\boldsymbol\beta}} + \mathbf{k}(\mathbf{x}^\prime, \mathbf{X})\mathbf{K}(\mathbf{X},\mathbf{X})^{-1}(\mathbf{y} -  \mathbf{h}(\mathbf{X})^{T}\mathbf{\hat{\boldsymbol{\beta}}})
\end{equation}
\begin{equation} \label{eq:predicted variance of explicit mean function}
    \hat{\sigma}^2(\mathbf{x}^\prime) = k(\mathbf{x}^\prime, \mathbf{x}^\prime)-\mathbf{k}\left( \mathbf{x}^\prime, \mathbf{X}\right )\mathbf{K}\left(\mathbf{X},\mathbf{X}\right)^{-1}\mathbf{k}\left(\mathbf{X}, \mathbf{x}^\prime \right)+ \mathbf{r}^{T}(\mathbf{h}(\mathbf{X})\mathbf{K}(\mathbf{X},\mathbf{X})^{-1}\mathbf{h}(\mathbf{X})^T)^{-1}\mathbf{r}
\end{equation}
where $ \mathbf{r}  = \mathbf{h}(\mathbf{x^\prime}) -\mathbf{h}(\mathbf{X})\mathbf{K}(\mathbf{X},\mathbf{X})^{-1}\mathbf{k} (\mathbf{X}, \mathbf{x^\prime})$. The first two
terms are predicted variance deduced by the zero mean GPR  and the third term is
the predicted variance caused by the explicit basis function.

Incorporating explicit basis functions helps the GPR to generalize better for more complicated problems, while one has to select the basis function $\mathbf{h(x)}$ beforehand or use symbolic regression to identify the terms \cite{Cranmer2023}.

\subsubsection{Parameters estimation}
\label{sec: params estimation of GPR}

Assuming the outputs $\mathbf{y}$ follow a Normal distribution, if a GPR with explicit basis functions is employed, the marginal likelihood can be formulated as \cite{Rasmussen2006}:

\begin{equation} \label{eq:marginal_likelihood}
    L \left( \mathbf{y} | \boldsymbol{\beta}, \boldsymbol{\theta} \right) = \frac{1}{\sqrt{(2\pi)^N |\mathbf{K}|}} \exp\left(-\frac{1}{2} (\mathbf{y}-\mathbf{h(X)}^T\boldsymbol{\beta})^T\mathbf{K}^{-1}(\mathbf{y}-\mathbf{h(X)}^T\boldsymbol{\beta})\right)
\end{equation}

Then, the log marginal likelihood can be expressed by:
\begin{equation} \label{eq: nll}
    \ln L \left( \mathbf{y} | \boldsymbol{\beta}, \boldsymbol{\theta} \right) =-\frac{N}{2}\ln(2\pi) - \frac{1}{2}\ln|\mathbf{K}| -\frac{(\mathbf{y}-\mathbf{h(X)}^T\boldsymbol{\beta})^T\mathbf{K}^{-1}(\mathbf{y}-\mathbf{h(X)}^T\boldsymbol{\beta})}{2}
\end{equation}

By taking derivatives of \Cref{eq: nll} and setting it to zero, we obtain
maximum likelihood estimation for $\boldsymbol{\beta}$.
\begin{equation} \label{eq: hat_beta}
    \hat{\boldsymbol\beta} = \left (\mathbf{h(X)}\mathbf{K}^{-1}\mathbf{h(X)}^T \right)^{-1}\mathbf{h(X)}\mathbf{K}^{-1}\mathbf{y}
\end{equation}

Now the maximum likelihood estimations $\hat{\boldsymbol{\beta}}$ can be substituted back into \Cref{eq: nll} and constant terms removed to give the concentrated ln-likelihood function:
\begin{equation} \label{eq: concentrated ln-likelihood}
    \ln L \left( \mathbf{y} | \boldsymbol{\theta} \right) =- \frac{1}{2}\ln|\mathbf{K}| -\frac{(\mathbf{y}-\mathbf{h(X)}^T\boldsymbol{\hat{\beta}})^T\mathbf{K}^{-1}(\mathbf{y}-\mathbf{h(X)}^T\boldsymbol{\hat{\beta}})}{2}
\end{equation}

The value of \Cref{eq: concentrated ln-likelihood} is determined by parameter $\bm{\theta}$, the maximum value can not be found analytically. Therefore,
heuristic algorithms can be utilized to find the optimum value of $\bm{\theta}$. 

\section{KRR-LR-GPR parameter estimation and prediction} \label{sec: paramete_estimation_of_mf_rbf_gp}

The parameter estimation of KRR-LR-GPR involves two stages, the first stage is to determine parameters for LF kernel regression, and the second is to tune parameters for the transfer-learning model and residual GPR. 

\subsection{LF-KRR parameter estimation} \label{sec: kernel regession}
KRR based on the LF dataset can be defined by 
\begin{equation} \label{eq: kernel regression}
    \mathbf{y}^l = \mathbf{K}(\mathbf{X}^l, \mathbf{X}^l) \mathbf{w}
\end{equation}

\noindent where $\mathbf{w}$ are the weights determined by LF data points, $\mathbf{K}(\mathbf{X}^l, \mathbf{X}^l)$ is a $N^l \times N^l$ Gram or covariance matrix with $\mathbf{K}_{ij} =k(\mathbf{x}^l_i, \mathbf{x}^l_j) $. The value of $\mathbf{w}$  can be determined analytically: 
\begin{equation} \label{eq: rbf regression}
     \mathbf{w} =\mathbf{K}( \mathbf{X}^l, \mathbf{X}^l)^{-1}\mathbf{y}^l
\end{equation}
If the data has noise, $\mathbf{w} = \left( \mathbf{K}(\mathbf{X}^l, \mathbf{X}^l )+ \lambda \mathbf{I}\right)^{-1}\mathbf{y}^l$ where $\lambda$ is the LF noise precision. 

\subsection{Transfer-learning model and residual GPR parameter estimation} 

Regarding the parameter estimation of the transfer-learning model and residual GPR, it follows the same procedure as  \Cref{sec: params estimation of GPR}, where the log marginal likelihood function can be formulated as: 

\begin{equation} \label{eq: nll_of_mf_rbf_gp}
    \ln L \left( \mathbf{y}^h | \boldsymbol{\rho}, \boldsymbol{\theta}^h \right) =-\frac{N^h}{2}\ln(2\pi) - \frac{1}{2}\ln|\mathbf{K}(\mathbf{X}^h, \mathbf{X}^h )| -\frac{(\mathbf{y}^h-\mathbf{m}(\mathbf{X}^h)^T\bm{\rho})^T\mathbf{K}(\mathbf{X}^h, \mathbf{X}^h )^{-1}(\mathbf{y}^h-\mathbf{m} (\mathbf{X}^h)^T\bm{\rho})}{2}
\end{equation}

By taking the derivative of \Cref{eq: nll_of_mf_rbf_gp} and setting it to zero, we can get the estimation of $\hat{\bm{\rho}}$ :
\begin{equation} 
    \label{eq: estimation_of_rho}
    \hat{\bm{\rho}}  = \left (\mathbf{m}(\mathbf{X}^h)\mathbf{K}(\mathbf{X}^h,\mathbf{X}^h )^{-1}\mathbf{m}(\mathbf{X}^h)^T \right)^{-1} \mathbf{m}(\mathbf{X}^h)\mathbf{K}(\mathbf{X}^h,\mathbf{X}^h )^{-1}\mathbf{y}^h
\end{equation}

Then, optimizing the ln-concentrated function for optimal $\bm{\theta}^h$

\begin{equation} \label{eq: concentrated ln-likelihood for proposed approach}
    \ln L \left( \mathbf{y}^h | \boldsymbol{\theta}^h \right) = - \frac{1}{2}\ln|\mathbf{K}(\mathbf{X}^h, \mathbf{X}^h )| -\frac{(\mathbf{y}^h-\mathbf{m}(\mathbf{X}^h)^T\boldsymbol{\hat{\rho}})^T\mathbf{K}(\mathbf{X}^h, \mathbf{X}^h )^{-1}(\mathbf{y}^h-\mathbf{m(X)}^T\boldsymbol{\hat{\rho}})}{2}
\end{equation}

\subsection{Predictions} 
We can obtain the predictions of KRR-LR-GPR by modifying \Cref{eq: gp with explicit mean} and (\ref{eq:predicted variance of explicit mean function})into: 
\begin{equation} \label{eq: predicted_mean_of_mf_rbf_gp}
    \hat{f}^h(\mathbf{x}^{\prime})=  \mathbf{m}(\mathbf{x^\prime})^{T}\mathbf{\hat{\bm\rho}} + \mathbf{k}(\mathbf{x}^\prime, \mathbf{X}^h)\mathbf{K}(\mathbf{X}^h,\mathbf{X}^h)^{-1}(\mathbf{y}^h -  \mathbf{m}(\mathbf{X}^h)^{T}\mathbf{\hat{\bm{\rho}}})
\end{equation}
\begin{equation} \label{eq:predicted_variance_of_mf_rbf_gpr}
    \hat{\sigma}_h^2(\mathbf{x}^\prime) = k(\mathbf{x}^\prime, \mathbf{x}^\prime)-\mathbf{k}\left( \mathbf{x}^\prime, \mathbf{X}^h\right )\mathbf{K}\left(\mathbf{X}^h,\mathbf{X}^h\right)^{-1}\mathbf{k}\left(\mathbf{X}^h, \mathbf{x}^\prime \right) + \mathbf{r}^{T}(\mathbf{m}(\mathbf{X}^h)\mathbf{K}(\mathbf{X}^h,\mathbf{X}^h)^{-1}\mathbf{m}(\mathbf{X}^h)^T)^{-1}\mathbf{r}
\end{equation}

where $ \mathbf{r}  = \mathbf{m}(\mathbf{x^\prime}) -\mathbf{m}(\mathbf{X}^h)\mathbf{K}(\mathbf{X}^h,\mathbf{X}^h)^{-1}\mathbf{k} (\mathbf{X}^h, \mathbf{x^\prime})$.

\section{Numerical functions}
\subsection{Numerical functions for testing KRR-LR-GPR}  \label{sec: low dimensional functions}

This section includes details on the low-dimensional numerical functions considered in this paper. Originally, those functions were single-fidelity functions used to
verify optimization algorithms \cite{simulationlib}. They were extended by
Jiang et al \cite{Jiang2019},  Sander et al. \cite{vanRijn2020} and Mainini et al. \cite{Mainini2022} for verifying the performance of MF optimization algorithms.

\begin{itemize}
    \item Forrester function
          \begin{align} \label{eq: forrester function}
              f^h(x) & = (6x - 2)^2 \sin(12x - 4)     \\
              f^l(x) & = A(6x - 2)^2 \sin(12x - 4) + B(x-0.5) - C \label{eq: forrester function2}
          \end{align}
          where  $x \in [0, 1]$. By selecting different values of $A$, $B$, and $C$, LF functions with different correlations to the HF function. 
    \item  Hartman3 function 
          \begin{align}
              f^h(\mathbf{x}) & = -\sum_{i=1}^{4} c_i \exp\left(-\sum_{j=1}^{3} a_{ij}(x_j - p_{ij})^2\right)          \\
              f^l(\mathbf{x}) & = 0.585 - 0.324x_1 - 0.379x_2 - 0.431x_3                                               \\
                              & \quad - 0.208x_1x_2 + 0.326x_1x_3 + 0.193x_2x_3 + 0.225x_1^2 + 0.263x_2^2 + 0.274x_3^2
          \end{align}

          where
          \[
              \mathbf{c} = \begin{bmatrix} 1 \\ 1.2 \\ 3 \\ 3.2 \end{bmatrix}, \quad
              \mathbf{p} = \begin{bmatrix} 0.3689 & 0.117 & 0.2673 \\ 0.4699 & 0.4387 & 0.747 \\ 0.1091 & 0.8732 & 0.5547 \\ 0.03815 & 0.5743 & 0.8828 \end{bmatrix}, \quad
              \mathbf{a} = \begin{bmatrix} 3 & 10 & 30 \\ 0.1 & 10 & 35 \\ 3 & 10 & 30 \\ 0.1 & 10 & 35 \end{bmatrix}
          \]
          where $\mathbf{x} \in [0, 1]$
    \item Hartman6 function
          \begin{align}
              f^h(\mathbf{x}) & = -\sum_{i=1}^{4} c_i \exp\left(-\sum_{j=1}^{6} a_{ij}(x_j - p_{ij})^2\right)     \\
              f^l(\mathbf{x}) & = -\sum_{i=1}^{4} c_i \exp\left(-\sum_{j=1}^{6} a_{ij}(l_j x_j - p_{ij})^2\right)
          \end{align}
          where
          \[
              \mathbf{p} = \begin{bmatrix}
                  0.1312 & 0.1696 & 0.5569 & 0.0124 & 0.8283 & 0.5886 \\
                  0.2329 & 0.4135 & 0.8307 & 0.3736 & 0.1004 & 0.9991 \\
                  0.2348 & 0.1451 & 0.3522 & 0.2883 & 0.3047 & 0.6650 \\
                  0.4047 & 0.8828 & 0.8732 & 0.5743 & 0.1091 & 0.0381
              \end{bmatrix},
              \quad
              \mathbf{c} = \begin{bmatrix} 1.0 \\ 1.2 \\ 3.0 \\ 3.2 \end{bmatrix},
          \]
          \[
              \mathbf{a} = \begin{bmatrix}
                  10.00 & 3.0  & 17.00 & 3.5  & 1.7  & 8  \\
                  0.05  & 10.0 & 17.00 & 0.1  & 8.0  & 14 \\
                  3.00  & 3.5  & 1.70  & 10.0 & 17.0 & 8  \\
                  17.00 & 8.0  & 0.05  & 10.0 & 0.1  & 14
              \end{bmatrix},
              \quad
              \mathbf{l} = \begin{bmatrix} 0.75 \\ 1.0 \\ 0.8 \\ 1.3 \\ 0.7 \\ 1.1 \end{bmatrix}
          \]
          where $\mathbf{x} \in [0, 1]$
    \item Six-hump function
          \begin{align}
              f^h(\mathbf{x}) & = (4 - 2.1x_1^2 + \frac{x_1^4}{3})x_1^2 + x_1x_2 - 4x_2^2 \\
              f^l(\mathbf{x}) & = f^h(0.7\mathbf{x}) - x_1x_2 - 15
          \end{align}
          where $\mathbf{x} \in [-2,2]$
    \item Bohachevsky function
          \begin{align}
              f^h(\mathbf{x}) & = (x_1^2 + 2x_2^2 - 0.3\cos(3\pi x_1) - 0.4\cos(4\pi x_2) + 0.7)^2 \\
              f^l(\mathbf{x}) & = f^h(0.7x_1, x_2) + x_1x_2-12
          \end{align}
          where $\mathbf{x} \in [ -5, 5]$
    \item Booth function
          \begin{align}
              f^h(\mathbf{x}) & = (x_1^2 + 2x_2^2 - 7)^2 + (2x_1^2 + x_2 - 5)^2               \\
              f^l(\mathbf{x}) & = f_h\left(0.4x_1, x_2 \right)+ 1.7 \cdot x_1x_2 - x_1 + 2x_2
          \end{align}
          where $\mathbf{x} \in [ -10, 10]$
    \item Borehole function
          \begin{align}
              f^b(\mathbf{x}, A, B) & = \frac{A T_u (H_u - H_l)}{\log\left(\frac{r}{r_w}\right) \left(1 + \frac{2L T_u}{\log\left(\frac{r}{r_w}\right) r_w^2 K_w} + \frac{T_u}{T_l}\right)} \\
              f^h(\mathbf{x})       & = f^b(\mathbf{x}, 2\pi, 1)                                                                                                                            \\
              f^l(\mathbf{x})       & = f^b(\mathbf{x}, 5, 1.5)
          \end{align}
          where

          \begin{align*}
              rw & \in [0.05, 0.15]    \\
              r  & \in [100, 50000]    \\
              Tu & \in [63070, 115600] \\
              Hu & \in [990, 1110]     \\
              Tl & \in [63.1, 116]     \\
              Hl & \in [700, 820]      \\
              L  & \in [1120, 1680]    \\
              Kw & \in [9855, 12045]
          \end{align*}

    \item CurrinExp function
          \begin{align}
              f^h(x_1, x_2) & = (1 - \exp(-1/(2x_2))) \times \frac{2300x_1^3 + 1900x_1^2 + 2092x_1 + 60}{100x_1^3 + 500x_1^2 + 4x_1 + 20} \\
              f^l(x_1, x_2) & = \frac{1}{4}\left[f^h(x_1+0.05, x_2+0.05) + f^h(x_1+0.05, x_2-0.05)\right] \nonumber                       \\
                            & + \frac{1}{4}\left[f^h(x_1-0.05, x_2+0.05) + f^h(x_1-0.05, x_2-0.05)\right]
          \end{align}
          where $\mathbf{x} \in [0, 1]$
          
    \item Park91A function
          \begin{align}
              f^h(\mathbf{x}) & = \frac{x_1}{2} \left(\sqrt{1 + (x_2 + x_3^2)\frac{x_4}{x_1^2}} - 1\right) + (x_1 + 3x_4)e^{1 + \sin(x_3)} \\
              f^l(\mathbf{x}) & = (1 + \sin(x_1)/10)f^h(\mathbf{x}) -2x_1 + x_2^2 +x_3^2 + 0.5
          \end{align}
          where $\mathbf{x} \in [0, 1]$
    \item Park91B function
          \begin{align}
              f^h(\mathbf{x}) & = \frac{2}{3}e^{x_1 + x_2} - x_4\sin(x_3) + x_3 \\
              f^l(\mathbf{x}) & = 1.2f^h(\mathbf{x}) - 1
          \end{align}
          where $\mathbf{x} \in [0, 1]$
\end{itemize}

\subsection{Numerical functions for testing DNN-LR-BNN}  \label{sec: high dimensional functions}
\begin{itemize}

\item 1D illustrative function \cite{Meng2021_MFBNN}
\begin{subequations}
\label{eq: meng_1d}
\begin{align}
    f^h(\mathbf{x}) & = (x-\sqrt{2})\sin\left(8\pi x\right)^2 \\
    f^{l1}(\mathbf{x}) & = \sin\left(8\pi x\right) \label{eq: meng_1d_lf1} \\
    f^{l2}(\mathbf{x}) & = 1.2 f^h(\mathbf{x})  -0.5  \label{eq: meng_1d_lf2} \\ 
    f^{l3}(\mathbf{x}) & = \sin\left(16\pi x\right)^2 \label{eq: meng_1d_lf3}
\end{align}
\end{subequations}
where $x \in [0, 1]$. In the illustrative case, both LF and HF data are assumed to be corrupted by white noise with $\mathcal{N}(0, 0.05^2)$.

\item 4D function \cite{Meng2021_MFBNN} 
\begin{align} \label{eq: Meng 4D}
    f^h(\mathbf{x}) & = \frac{1}{2} (0.1 \exp(x_1 + x_2) - x_4 \sin(12\pi x_3) + x_3) \\
    f^l(\mathbf{x}) & = 1.2 f^h(\mathbf{x}) -0.5
\end{align}

where $x \in [0, 1]$. The LF has noise with $\mathcal{N}(0, 0.05^2)$ and HF has noise with $\mathcal{N}(0, 0.01^2)$

\item  high dimensional function   

The high dimensional function used in \cite{Meng2020_MFPINN} is adopted, which can be expressed by
\begin{align} \label{eq: Meng 20}
    f^h(\mathbf{x}) & = \sum_{i=2}^{20} \left(2x_i^2 - x_{i-1} \right)^2 + (x_1-1)^2 \\
    f^l(\mathbf{x}) & = 0.8f^h(\mathbf{x}) + \sum_{i=2}^{20} 0.4x_{i-1}x_i -50
\end{align}

where $x \in [-3, 3]$. We set the dimension in [20, 50, 100] to challenge DNN-LR-BNN in high-dimensional problems. Moreover, this function is assumed to have zero noise in LF and HF has noise with $\mathcal{N}(0, 50^2)$.

\end{itemize}

\section{Additional experiments for KRR-LR-GPR} \label{sec:additional experiments for KRR-LR-GPR} 

\subsection{Additional results of KRR-LR-GPR comprehensive experiments} \label{sec: additional gpr experiments results}

\begin{figure}
    \centering
    \includegraphics[width=\textwidth]{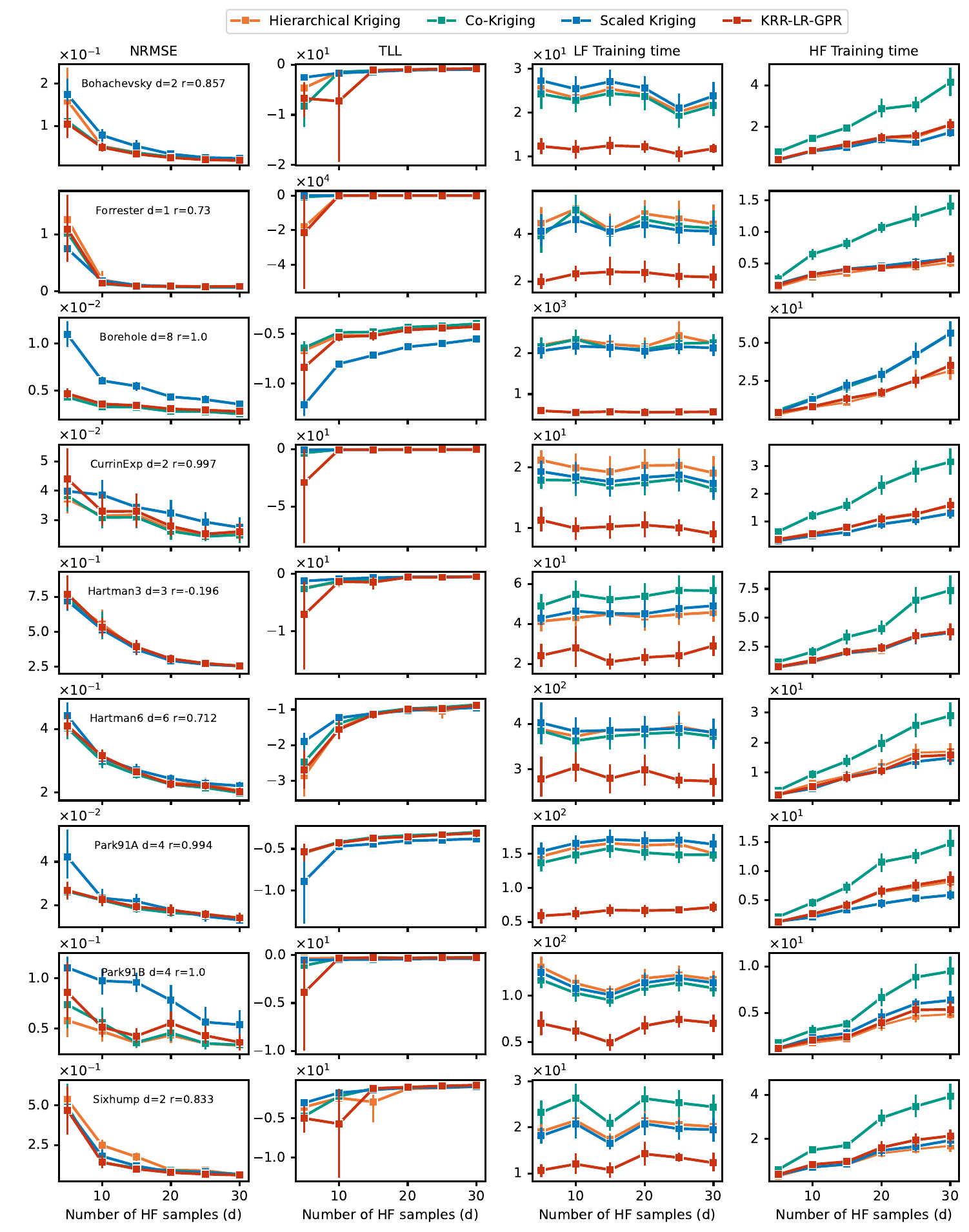}
    \caption{Additional comprehensive experiment results of KRR-LR-GPR with different amounts of HF data.  Different colors represent different methods, where NRMSE is the smaller the better and TLL is the larger the better.}\label{fig:additional_comprehensive_experiments_200lf}
\end{figure}

\begin{figure}
    \centering
    \includegraphics[width=\textwidth]{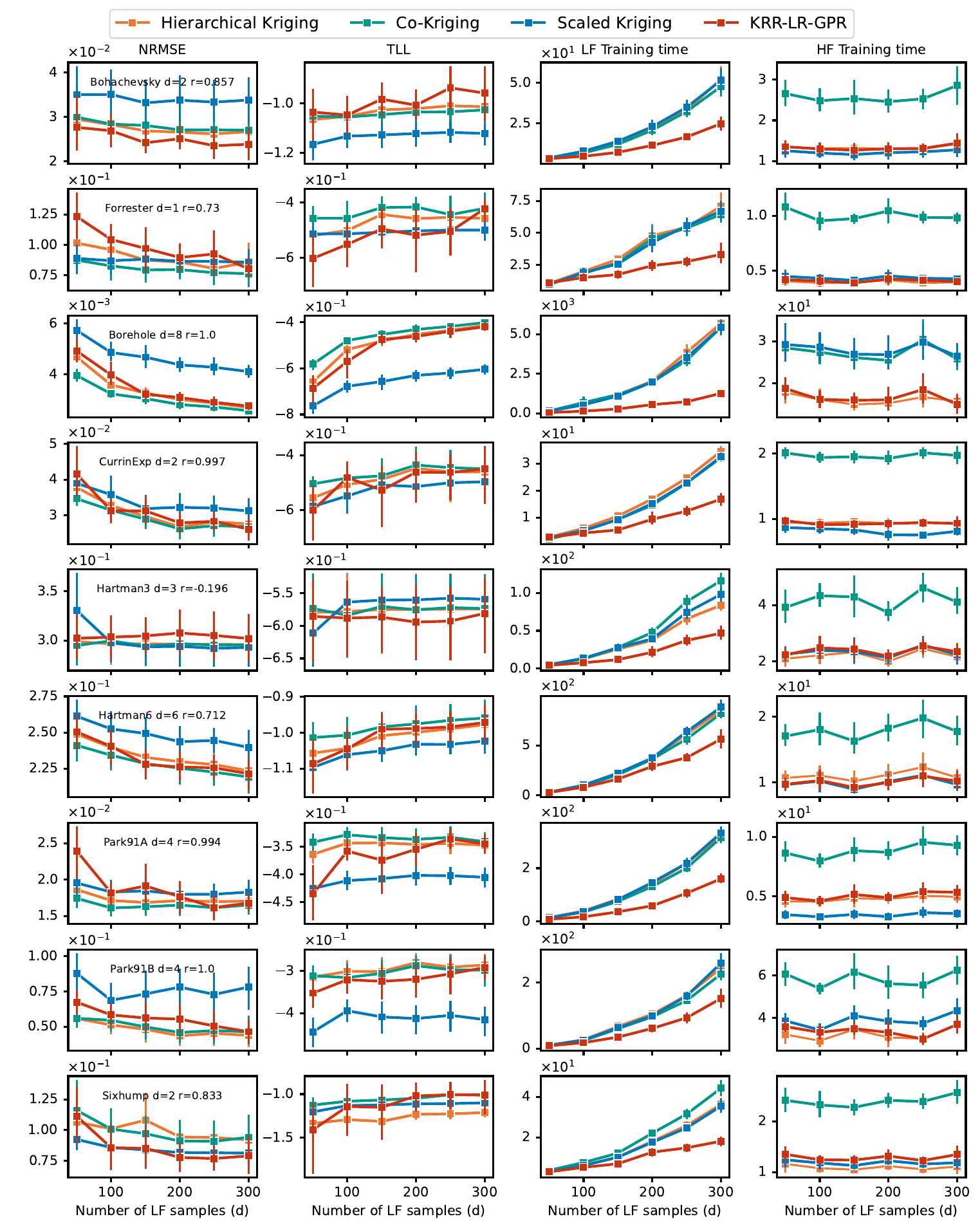}
    \caption{Additional comprehensive experiment results of KRR-LR-GPR with different amounts of LF data.  Different colors represent different methods, where NRMSE is the smaller the better and TLL is the larger the better.}
    \label{fig:additional_comprehensive_experiments_20hf}
\end{figure}

In \Cref{sec: comprehensive experiments of KRR-LR-GPR}, we only show the results of the \textit{Booth} function, and the remaining comparison results are shown in \Cref{fig:additional_comprehensive_experiments_200lf} and  \Cref{fig:additional_comprehensive_experiments_20hf}. 

It can be observed that the proposed KRR-LR-GPR method achieves performance comparable to other state-of-the-art approaches in both scenarios presented in  \Cref{fig:additional_comprehensive_experiments_200lf} and  \Cref{fig:additional_comprehensive_experiments_20hf}. Moreover, its training time for LF data is significantly lower than that of other methods. These results highlight the potential of KRR-LR-GPR in advancing data-scarce modeling by leveraging larger LF datasets, particularly in machine learning modeling scenarios where computational cost is dominated by LF data. On the other hand, when both LF and HF data are limited, KRR-LR-GPR remains a reliable and competitive choice as demonstrated in \Cref{sec:naca0012_airfoil}.

\subsection{Experiments with noiseless data-scarce HF and low-dimensional problems: KRR-LR-GPR} \label{sec: noiseless experiment results}

\begin{figure}[h]
    \centering
    \includegraphics[width = \textwidth]{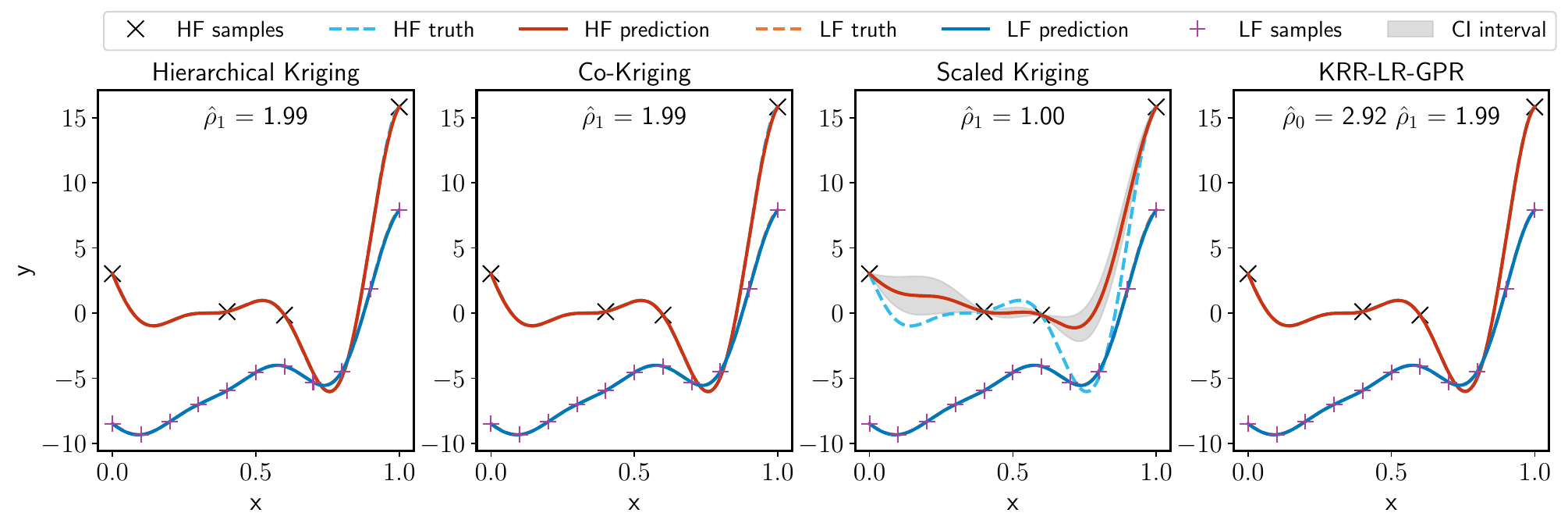}
    \caption{Fitting performance of MF models on noiseless illustrative example under 5 independent runs of random seeds}
    \label{fig:1d illustrative example of noiseless mf_rbf_kriging}
\end{figure}

As reported in the literature \cite{Giannoukou2024} and to the best of our knowledge, the existing MF-GPR methods have been considered for noiseless problems, which is a special case in this paper where $\sigma_h = 0$ and $\sigma_l = 0$. To demonstrate the effectiveness of the KRR-LR-GPR in those cases, we present the illustrative example based on the typical MF case presented by Forrester \cite{Han2012}, the predicted performance of KRR-LR-GPR compared to other MF-GPR approaches is presented in \Cref{fig:1d illustrative example of noiseless mf_rbf_kriging} and different performance metrics for this problem are included in  \Cref{tab:results_1D_noiseless_mfrbfgpr}.

\begin{table}[h]
    \centering
    \caption{Results comparison of the noiseless illustrative example with 5 independent runs}
    \label{tab:results_1D_noiseless_mfrbfgpr}
    \begin{tabular}{c m{2.5cm} m{2.5cm} m{2.5cm} m{2.5cm}}
        \hline
        \multirow{2}{*}{Methods}                     &\multirow{2}{*}{NRMSE}            & \multirow{2}{*}{R2 Score}        &  \multicolumn{2}{c}{Training time ($s$)} \\ 
        & & &  LF  &  HF \\
        \hline
        Hierarchical Kriging          & 0.0228  \tiny{($\pm 1.49 \times 10^{-5}$)}        & 0.9998 \tiny{($\pm 2.07 \times 10^{-7}$)}        & 0.2673 \tiny{($\pm 6.05 \times 10^{-2}$)}& \textbf{0.0426} \tiny{($\pm 6.67\times 10^{-3}$)} \\
        Co-Kriging               & 0.0228 \tiny{($\pm 1.49 \times 10^{-5}$)}         & 0.9998 \tiny{($\pm 2.06 \times 10^{-7}$)}        & 0.2963 \tiny{($\pm 3.80 \times 10^{-2}$)}  & 0.4214 \tiny{($\pm 7.25 \times 10^{-2}$)}            \\
        Scaled Kriging            & 0.9754  \tiny{($\pm 1.78 \times 10^{-4}$)}        & 0.6792 \tiny{($\pm 6.99 \times 10^{-5}$)}         & 0.2934  \tiny{($\pm 3.54 \times 10^{-2}$)} & 0.0886  \tiny{($\pm 4.29 \times 10^{-4}$)}      \\
        KRR-LR-GPR     & \textbf{0.0072} \tiny{($\pm 2.44 \times 10^{-6}$)} & \textbf{1.0000} \tiny{($\pm 1.00 \times 10^{-8}$)} & \textbf{0.0003} \tiny{($\pm 5.51 \times 10^{-5}$)} & 0.0569 \tiny{($\pm 9.00 \times 10^{-3}$)} \\
        \hline
    \end{tabular}
\end{table}

We can observe that the findings are the same as the results reported in  \Cref{fig:1d_noisy_gpr_illustration_case} and \Cref{tab:results_1D_noisy_mfrbfgpr}. KRR-LR-GPR has a slightly better performance compared with other MF approaches. Meanwhile, we can see that KRR-LR-GPR is faster to train compared with other methods.

\section{Ablation studies of KRR-LR-GPR} \label{sec: abaltion studies}

\subsection{Influence of LF polynomial order to KRR-LR-GPR} \label{sec: lf basis selection}

\begin{figure}[hbt!]
    \centering
    \includegraphics[width=\textwidth]{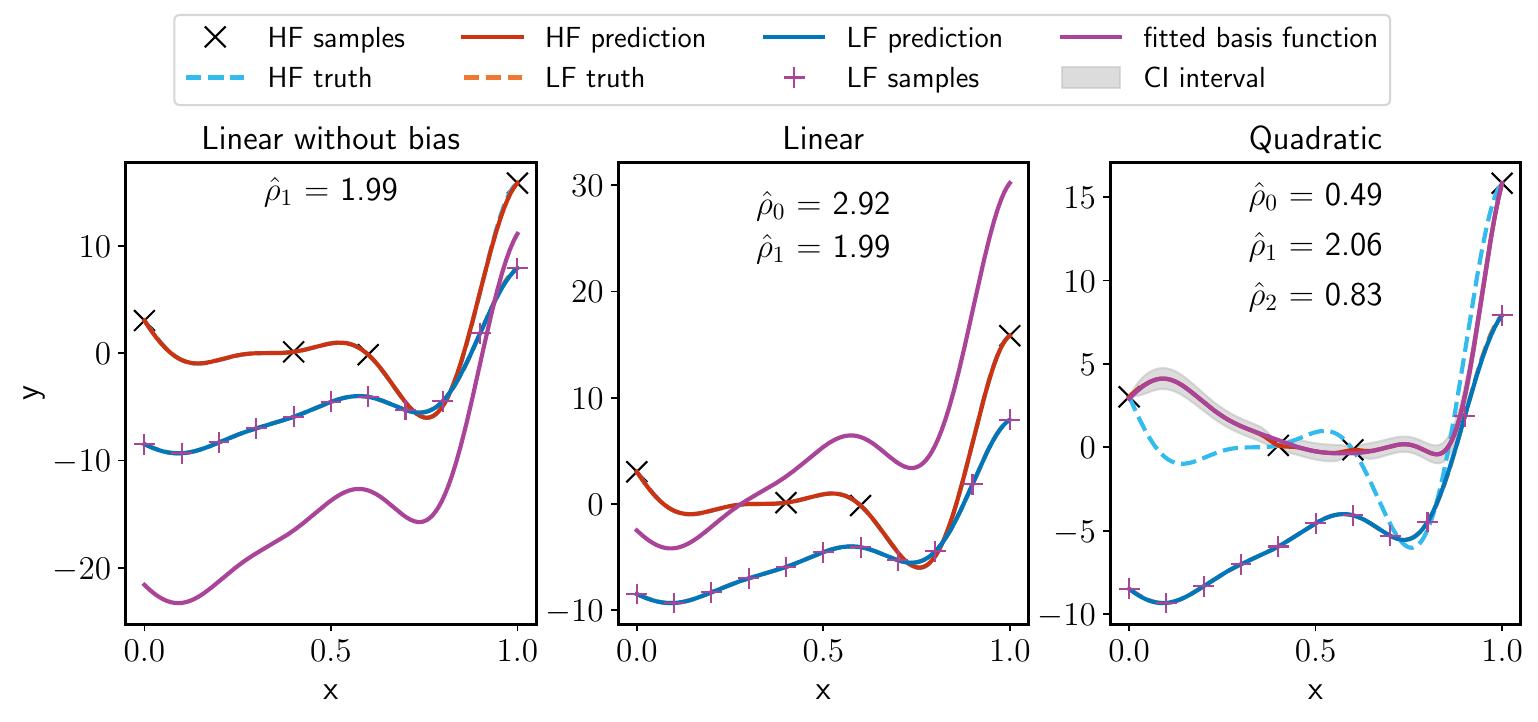}
    \caption{Performance of KRR-LR-GPR based on different LF polynomial orders}
    \label{fig: lf_polynomial_orders}
\end{figure}

We provide an opportunity to select different polynomial orders of the LF surrogate in \Cref{eq: mf-bml}) and  \Cref{fig:mf_bml_framework}. We conducted an ablation study on how the selection of $\mathbf{m}(\mathbf{x})$ would influence the performance of the KRR-LR-GPR\footnote{To eliminate the influence of noise, the experiment is conducted with datasets without noise.}. \Cref{fig: lf_polynomial_orders} shows the fitting performance on an illustrative example, and \Cref{tab: different_lf_basis_polynomial_order} summarizes the corresponding accuracy metric values.

\begin{table}[h]
    \centering
    \caption{Results comparison of KRR-LR-GPR based on different LF polynomial orders}
    \label{tab: different_lf_basis_polynomial_order}
    \begin{tabular}{c m{2.5cm} m{2.5cm} m{2.5cm} m{2.5cm}}
        \hline
        \multirow{2}{*}{Methods}                     &\multirow{2}{*}{NRMSE}            & \multirow{2}{*}{R2 Score}        &  \multicolumn{2}{c}{Training time ($s$)} \\ 
        & & &  LF  &  HF \\
        \hline
        Linear without bias       & 0.0073   \tiny{($\pm 2.55 \times 10^{-6}$)}         & 0.9999  \tiny{($\pm 9.74 \times 10^{-9}$)}    & 0.0002  \tiny{($\pm 2.19\times 10^{-4}$)}  & 0.0417  \tiny{($\pm 5.13 \times 10^{-3}$)}  \\
        Linear               & \textbf{0.0072}  \tiny{($\pm 2.55 \times 10^{-6}$)}         &  \textbf{1.0000}   \tiny{($\pm9.76 \times 10^{-9}$)}       &   0.0002  \tiny{($\pm 7.45 \times 10^{-5}$)}   & 0.0342  \tiny{($\pm 8.93 \times 10^{-3}$)}  \\
        Quadratic       & 1.1358  \tiny{($\pm 1.79 \times 10^{-4}$)}         &0.5651  \tiny{($\pm 8.73 \times 10^{-5}$)}        &   0.0002  \tiny{($\pm 3.11 \times 10^{-5}$)}  & 0.0450  \tiny{($\pm 1.11 \times 10^{-2 }$)}   \\
        \hline
    \end{tabular}
\end{table}

\begin{figure}[h]
    \centering
    \includegraphics[width=\textwidth]{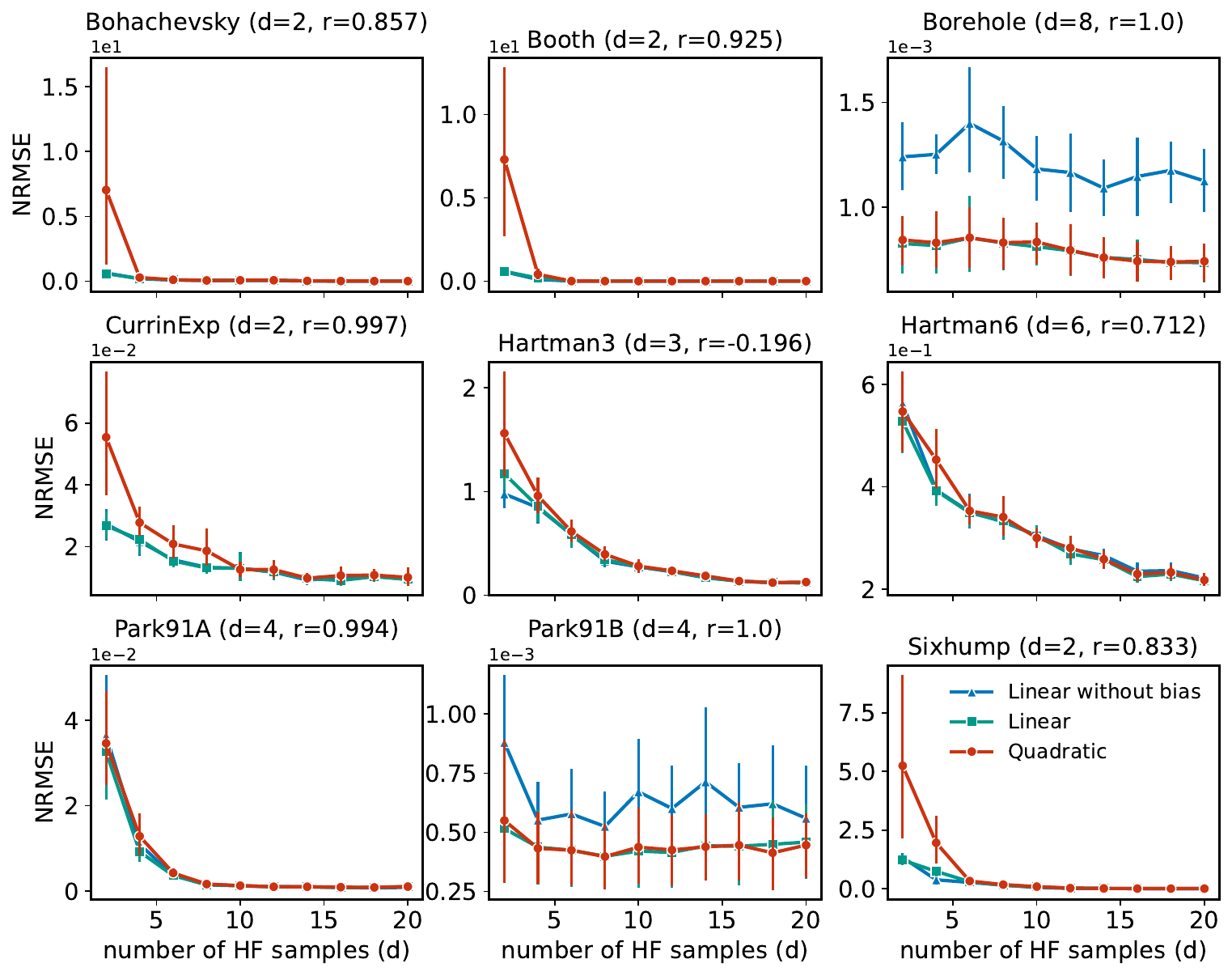}
    \caption{NRMSE of MF-RBF-Kriging based on different LF polynomial orders (LF data $100 \times d$)}
    \label{fig: comprehensive_experiments_on_lf_poly_order}
\end{figure}

With quadratic LF features, the KRR-LR-GPR overfits the Forrester function, though $\mathbf{m}(\mathbf{x})\bm{\rho}$ have a small residual to $\mathbf{y}^h$. On the other hand, KRR-LR-GPR  can fit the Forrester function well with the linear LF feature, where we observe that the bias does influence the value of $\rho_1$, but shifts the entire LF surrogate closer to the HF data. As listed in \Cref{tab: different_lf_basis_polynomial_order} the bias term $\rho_0$ also improves accuracy. To gain a deeper understanding of the effect of the LF polynomial order,  \Cref{fig: comprehensive_experiments_on_lf_poly_order} gives the NRMSE changes as the HF sample increases on the set of test examples.

As shown in \Cref{fig: comprehensive_experiments_on_lf_poly_order}, KRR-LR-GPR with quadratic LF polynomial order has large errors, leading to overfitting in most cases when the number of LF samples is smaller. The cases with linear polynomial order alleviate this issue to some extent. Regarding the effect of the bias term $\rho_0$, it does not influence most of the cases, whereas it brings improvements in some cases like \textit{Borehole} and \textit{Park91A}. Moreover, the bias term has slight positive effects on KRR-LR-GPR when LF samples are less in most cases. To this end, we adopt $\mathbf{m}(\mathbf{x}) = [1, f^l(\mathbf{x})]^T$ for KRR-LR-GPR in this paper. 

\subsection{Influence of HF-to-LF correlation and HF noise level on KRR-LR-GPR}  \label{HF-to-LF correlation}

\begin{figure}[h]
    \centering
    \includegraphics[width=\textwidth]{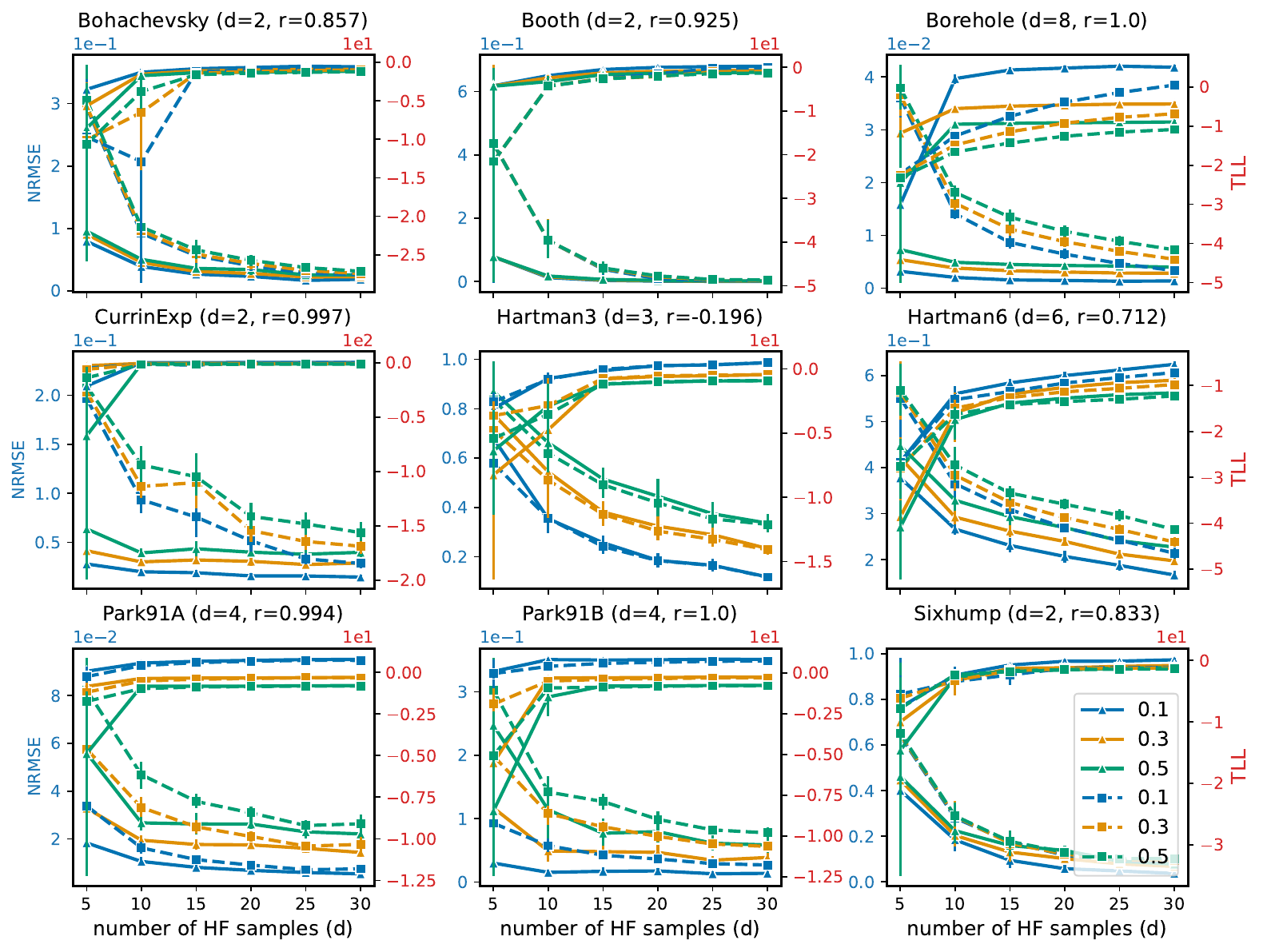}
    \caption{NRMSE and TLL of different examples with increasing HF data (LF data $200 \times d$): The solid and dash lines represent KRR-LR-GPR and GPR respectively, and line color shows different HF noise levels}
    \label{fig: mf_rbf_kriging_correlation_influence}
\end{figure}

KRR-LR-GPR converges to good accuracy with more HF data as shown in \Cref{sec: comprehensive experiments of KRR-LR-GPR}, while differences exist among different examples due to the HF-to-LF correlation, see \Cref{sec: additional gpr experiments results}. We fix the noise level to be $\sigma_h = 0.3$ in \Cref{sec: comprehensive experiments of KRR-LR-GPR}. Therefore, we design the ablation study here to investigate the influence of the HF-to-LF correlation and the HF noise level on KRR-LR-GPR compared to GPR. \Cref{fig: mf_rbf_kriging_correlation_influence} depicts the corresponding results.

We can see from \Cref{fig: mf_rbf_kriging_correlation_influence}  that the HF noise level has a significant impact on KRR-LR-GPR, where larger HF noise levels would deteriorate the performance even with an adequate amount of HF data. Specifically, we provide three noise levels for each test function that are $\sigma_h \in [0.1, 0.3, 0.5]$. It is clear to see that the accuracy metrics are better in the case where the noise is smaller for each test function. In addition, higher HF-to-LF correlations are preferred when utilizing KRR-LR-GPR, where clear advantages are observed on both NRMSE and TLL such as in the cases of \textit{Bohachevsky}, \textit{Booth}, \textit{Borehole}, \textit{CurrinExp}, \textit{Hartman6}, \textit{Park91A}, \textit{Park91B}, and the \textit{Sixhump} functions. Conversely, poor HF-to-LF correlation decreases the performance of KRR-LR-GPR, for instance, there is margin improvement over GPR in the case of \textit{Hartman3} function where the Pearson correlation coefficient between HF and LF is $-0.196$.

\subsection{Influence of HF and LF samples on KRR-LR-GPR} 
\label{sec: ablation study of KRR-LR-GPRR}

\begin{figure}[h]
    \centering
    \includegraphics[width=\textwidth]{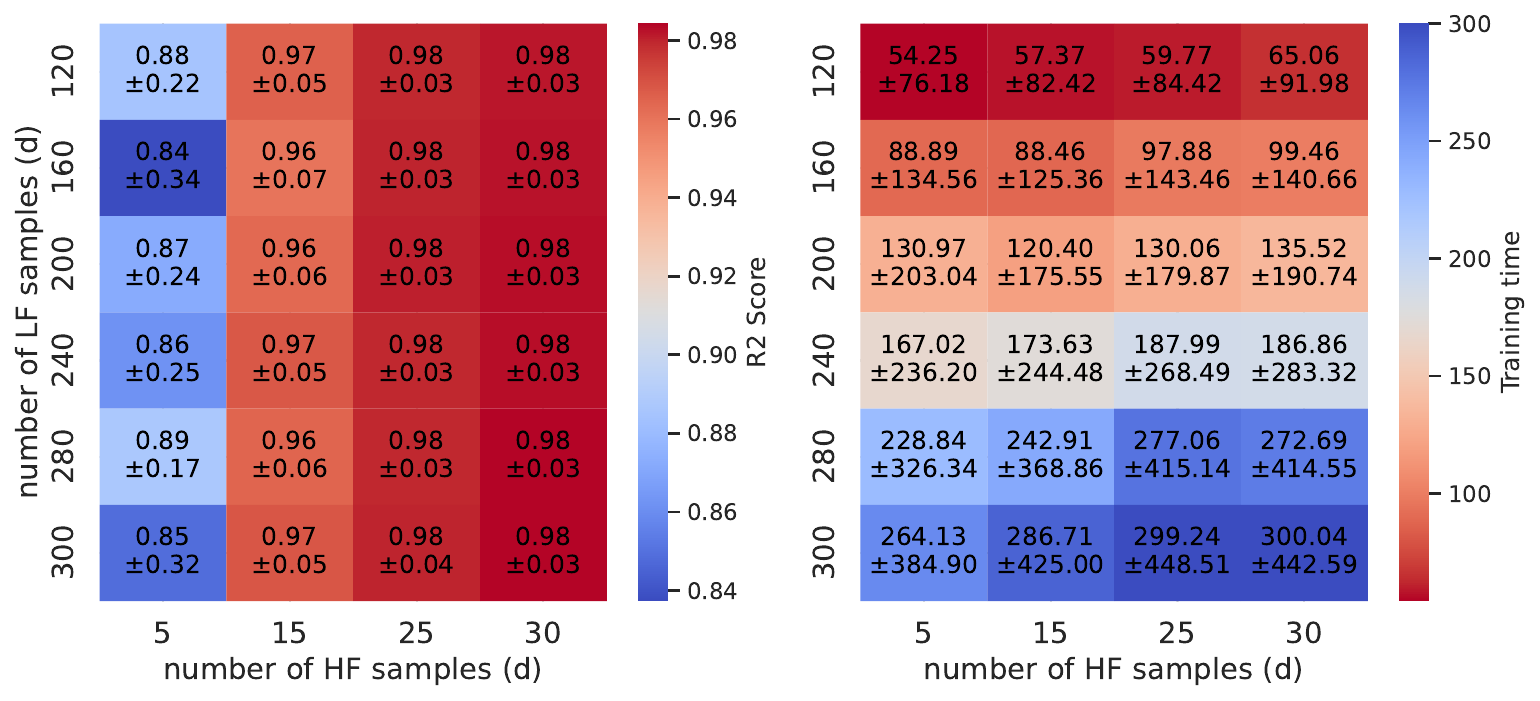}
    \caption{Influence of HF and LF samples on KRR-LR-GPR. The heatmap aggregates all test functions listed in Section \ref{sec: low dimensional functions} and every experiment is repeated 5 times by resampling from the design space independently}
    \label{fig:mf_rbf_gpr_sample_influence}
\end{figure}

\Cref{sec: comprehensive experiments of KRR-LR-GPR} included two experiments using a fixed number of HF or LF samples and explored the effect of another factor. To have a better overview of how HF and LF samples influence the KRR-LR-GPR method, \Cref{fig:mf_rbf_gpr_sample_influence} \footnote{R2 score is selected because it does not depend on the actual magnitude of the response.} shows the whole design of experiments instead of slices shown in \Cref{fig: gpr_comparison_at_200lf_samples} and \ref{fig:gpr_comparison_at_20hf_samples}. 

In \Cref{fig:mf_rbf_gpr_sample_influence}, the R2 Score is primarily influenced by the number of HF samples, and the LF samples will improve the robustness of KRR-LR-GPR. The training time plotted at the right heatmap is the total training time for both HF and LF models, more CPU time is needed with a greater total amount of data. However, the influence of HF data on execution time is more critical than that of LF data. 

\section{Hyperparameters of experiments and additional experiments on DNN-LR-BNN} \label{sec: hyperparameters of DNN-LR-BNN}

\subsection{Hyperparameters of the higher dimensional experiments}
Hyperparameters and sampling points used for training the DNN-LR-BNN and DNN-BNN models are described in  \Cref{tab:hyperparams}. These experiments were extracted from references \cite{Meng2020_MFPINN, Meng2021_MFBNN, Guo2022} for examples of 1D, 4D, and 20D. For the cases of 50D and 100D, we adopt the same hyperparameters as for the case of 20D and consider $1000\times d$ and $100\times d$ samples for LF and HF, respectively.

\begin{table}[h]
    \centering
    \caption{Hyperparameters for the DNN-LR-BNN experiments}
    \label{tab:hyperparams}
    \begin{tabular}{c c c >{\centering\arraybackslash}p{2.5cm}>{\centering\arraybackslash}p{2.5cm}>{\centering\arraybackslash}p{2.5cm} p{1.5cm}}
        \hline
        \multirow{2}{*}{Example} & \multicolumn{2}{c}{\multirow{2}{*}{Hyperparameter Name}} & \multicolumn{3}{c}{Hyperparameters setting} & \multirow{2}{*}{\parbox{3cm}{Samples}} \\
        \cline{4-6}
         & \multicolumn{2}{c}{} & BNN & DNN-BNN & DNN-LR-BNN & \\
        \hline
        \multirow{12}{*}{1D} 
        & \multirow{6}{*}{DNN} 
        & Architecture & - & 2 hidden, 50 neurons, \textit{Tanh} & 2 hidden, 50 neurons, \textit{Tanh} & \multirow{12}{*}{\begin{tabular}{@{}c@{}}201 LF \\ 11 HF \end{tabular}} \\
        & & Learning Rate & - &  $10^{-3}$ &  $10^{-3}$ & \\
        & & Optimizer & -  &  Adam &  Adam & \\
        & & Epoch & - & 10000  & 10000 & \\
        \cline{2-6}
        & \multirow{6}{*}{BNN} 
        &  Architecture & 2 hidden, 512 neurons, \textit{Tanh} & 2 hidden, 50 neurons, \textit{Tanh} & 2 hidden, 512 neurons, \textit{Tanh} & \\
        & & Learning rate & 0.001 & 0.001 & 0.001 & \\
        & & Posterior collection & 300 samples, Burn-in: 20000, Frequency: 100 & 300 samples, Burn-in: 20000, Frequency: 100 & 300 samples, Burn-in: 20000, Frequency: 100 & \\
        \hline
        \multirow{12}{*}{4D} 
        & \multirow{6}{*}{DNN} 
        & Architecture & - & 2 hidden, 256 neurons, \textit{Tanh} & 2 hidden, 256 neurons, \textit{Tanh} & \multirow{12}{*}{\begin{tabular}{@{}c@{}}25000 LF \\ 150 HF \end{tabular}} \\
        & & Learning Rate & - & $10^{-3}$ & $10^{-3}$ & \\
        & & Optimizer & - & Adam& Adam & \\
        & & Epochs & -  & 50000  & 50000 & \\
        \cline{2-6}
        & \multirow{6}{*}{BNN} 
        & Architecture & 2 hidden, 50 neurons , \textit{Tanh}   &  2 hidden, 50 neurons , \textit{Tanh} & 2 hidden, 50 neurons , \textit{Tanh}  & \\
        & & Learning Rate & $10^{-3}$  & $10^{-3}$ & $10^{-3}$ & \\
        & & Posterior Collection & 300 samples, Burn-in: 20000, Frequency: 100 & 300 samples, Burn-in: 20000, Frequency: 100 & 300 samples, Burn-in: 20000, Frequency: 100 & \\
        \hline
        \multirow{12}{*}{20D} 
        & \multirow{6}{*}{DNN} 
        & Architecture & - & 2 hidden, 200 neurons, \textit{Tanh} & 2 hidden, 200 neurons, \textit{Tanh} & \multirow{12}{*}{\begin{tabular}{@{}c@{}}30000 LF \\ 5000 HF \end{tabular}} \\
        & & Learning Rate & - & $10^{-3}$ & $10^{-3}$ & \\
        & & Optimizer & - & Adam& Adam & \\
        & & Epochs & -  & 80000  & 80000 & \\
        \cline{2-6}
        & \multirow{6}{*}{BNN} 
        & Architecture & 2 hidden, 512 neurons , \textit{ReLU}   &  2 hidden, 512 neurons , \textit{ReLU} & 2 hidden, 512 neurons , \textit{ReLU}  & \\
        & & Learning Rate & $10^{-3}$  & $10^{-3}$ & $10^{-3}$ & \\
        & & Posterior Collection & 300 samples, Burn-in: 20000, Frequency: 100 & 300 samples, Burn-in: 20000, Frequency: 100 & 300 samples, Burn-in: 20000, Frequency: 100 & \\
        \hline
        \multirow{12}{*}{50D} 
        & \multirow{6}{*}{DNN} 
        & Architecture & - & 2 hidden, 256 neurons, \textit{Tanh} & 2 hidden, 256 neurons, \textit{Tanh} & \multirow{12}{*}{\begin{tabular}{@{}c@{}}50000 LF \\ 5000 HF \end{tabular}} \\
        & & Learning Rate & - & $10^{-3}$ & $10^{-3}$ & \\
        & & Optimizer & - & Adam& Adam & \\
        & & Epochs & -  & 50000  & 50000 & \\
        \cline{2-6}
        & \multirow{6}{*}{BNN} 
        & Architecture & 2 hidden, 512 neurons , \textit{ReLU}   &  2 hidden, 512 neurons , \textit{ReLU} & 2 hidden, 512 neurons , \textit{ReLU}  & \\
        & & Learning Rate & $10^{-3}$  & $10^{-3}$ & $10^{-3}$ & \\
        & & Posterior Collection & 400 samples, Burn-in: 10000, Frequency: 100 & 400 samples, Burn-in: 10000, Frequency: 100 & 400 samples, Burn-in: 10000, Frequency: 100 & \\
        \hline
        \multirow{12}{*}{100D} 
        & \multirow{6}{*}{DNN} 
        & Architecture & - & 2 hidden, 256 neurons, \textit{Tanh} & 2 hidden, 256 neurons, \textit{Tanh} & \multirow{12}{*}{\begin{tabular}{@{}c@{}}100000 LF \\ 10000 HF \end{tabular}} \\
        & & Learning Rate & - & $10^{-3}$ & $10^{-3}$ & \\
        & & Optimizer & - & Adam& Adam & \\
        & & Epochs & -  & 50000  & 50000 & \\
        \cline{2-6}
        & \multirow{6}{*}{BNN} 
        & Architecture & 2 hidden, 512 neurons , \textit{ReLU}   &  2 hidden, 512 neurons , \textit{ReLU} & 2 hidden, 512 neurons , \textit{ReLU}  & \\
        & & Learning Rate & $10^{-3}$  & $10^{-3}$ & $10^{-3}$ & \\
        & & Posterior Collection & 400 samples, Burn-in: 10000, Frequency: 100 & 300 samples, Burn-in: 10000, Frequency: 100 & 400 samples, Burn-in: 10000, Frequency: 100 & \\
        \hline
    \end{tabular}
\end{table}

\subsection{Hyperparameters of the material structure-property linkages problem}
\label{sec:hyper_parms_mat_law}

The hyperparameters for the material structure-property linkages problem, as described in \Cref{sec:mat_law_prediction}, are detailed below.  For the DNN settings, both DNN-BNN and DNN-LR-BNN use an architecture with two hidden layers of 256 neurons and ReLU activation. The learning rate is set to $10^{-3}$, and the optimizer used is Adam. Both models are trained for 5000 epochs. For the BNN settings, BNN employs an architecture with two hidden layers of 256 neurons and ReLU activation, The learning rate for all models is 0.001. Posterior collection involves 100 data points with a burn-in of 10,000 iterations and a frequency of 100. The results presented in \Cref{sec:additional_results_of_mat_law_prediction} follow the same settings, except that BNN uses a two-layer architecture with 100 neurons per layer.

The dataset consists of 3200 LF data points and 500 HF data points. For single-fidelity BNN training, 400 HF data points are utilized;  and the MF methods employ 3200 LF and 100 HF data points. We claim that both single-fidelity and MF methods have the same computational cost for data generation in total. 

\begin{figure}[h]
    \centering
    \includegraphics[width=\textwidth]{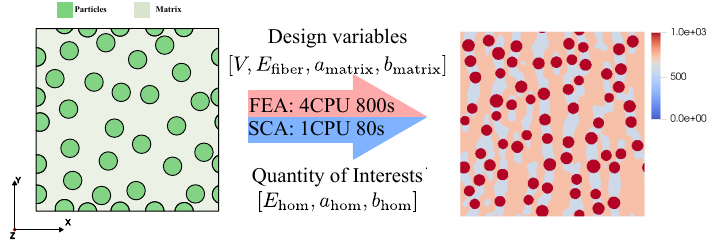}
    \caption{Illustration of the simulation process for the 2-phase material SVE. The left figure presents a schematic representation of a 2-phase microstructure consisting of matrix and particle materials. For any given set of input variables, the simulation can be performed using either FEA or SCA, yielding the corresponding strain and stress fields.  In the right figure, we depict the stress field for an arbitrary input: \( V = 0.2 \), \( E_{\text{fiber}} = 400000 \), \( a_{\text{matrix}} = 400 \), and \( b_{\text{matrix}} = 0.3 \). Based on the simulation results, we post-process the strain and stress fields to extract the QoIs, with further details provided in \Cref{fig: illustration of homoginization}.
}
    \label{fig: sve simulation illustration}
\end{figure}

\begin{figure}[h]
    \centering
    \includegraphics[width=0.9\textwidth]{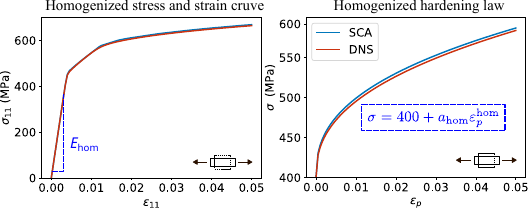}
    \caption{Comparison between the homogenized properties of FEA and SCA. The left figure shows the homogenized strain and stress curve along the loading direction(11 direction from Mechanics of Material) and from this curve, we can obtain $E_\text{hom}$ which is depicted slope. The right figure gives the homogenized hardening law which describes the relation between accumulated plastic strain $\varepsilon_p$ and Von Mises stress $\sigma$. From which, we can get $a_\text{hom}$ and $b_\text{hom}$ with a simple curve fit.}
    \label{fig: illustration of homoginization}
\end{figure}

Regarding the simulation setups, the design variables given in \Cref{sec:mat_law_prediction} which are the $[V, E_\text{fiber}, a_\text{matrix}, b_\text{matrix}]$. Other simulation parameters include a Poisson’s ratio of 0.25 for the elastic fiber and 0.30 for the plastic matrix. Young's modulus of the matrix is set to $100000$ MPa, and the yield stress in the matrix hardening law is $400$ MPa. The SVE is discretized into 300 elements per edge, resulting in 90,000 structural elements. We give an illustration of the simulation in \Cref{fig: sve simulation illustration} where the design variables have been passed into the simulation and corresponding strain and stress fields could be obtained through FEA or SCA. Based on the stain/stress fields of the simulation. we then post-process and obtain the homogenized QoIs as shown in \Cref{fig: illustration of homoginization}.

As shown in \Cref{fig: sve simulation illustration} and \ref{fig: illustration of homoginization}, SCA serves as an effective simplification, with both curves in \Cref{fig: illustration of homoginization} closely matching those obtained from FEA. Moreover, SCA significantly accelerates the simulation, reducing CPU execution time to approximately $\frac{1}{32}$ of that required by FEA.

\subsection{Influence of neural architecture on the performance of DNN-LR-BNN} \label{sec:hyperparameter-study}

\subsubsection{Illustrative example}

\begin{figure}[h]
    \centering
    \includegraphics[width=\textwidth]{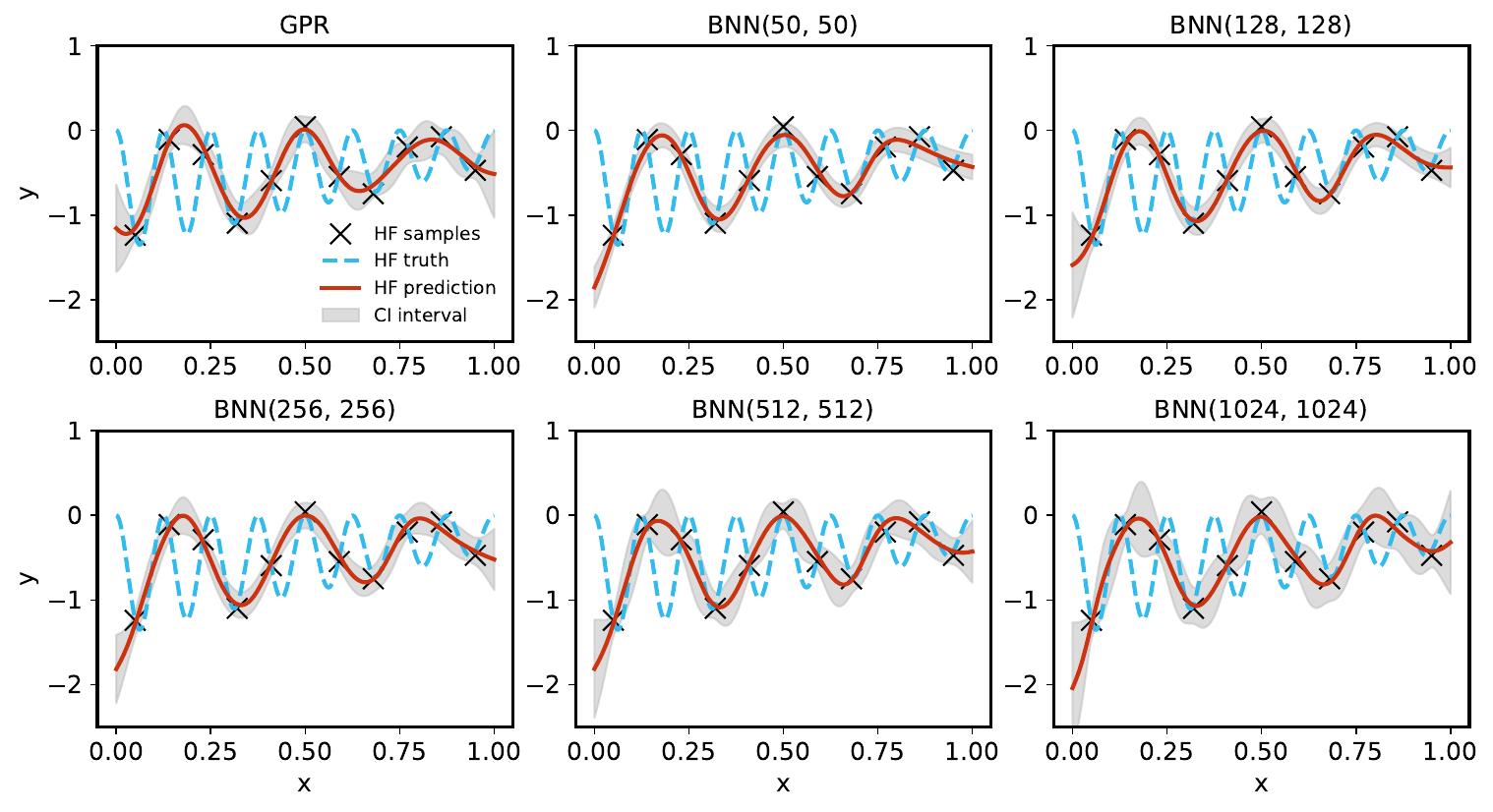}
    \caption{Influence of network architecture on the performance of single-fidelity BNN. All the hyperparameters are fixed as  \Cref{tab:hyperparams} except expanding the network from (50, 50) to (1024, 1024) neurons in each of the two hidden layers.}
    \label{fig:bnn_network_architecture}
\end{figure}

We know that hyperparameters are crucial to DNNs. Therefore, we conduct a brief study on the influence of neural architecture on the performance of the developed DNN-LR-BNN method based on the illustrative example in \Cref{sec: 1d_mf_dnn_bnn}. First of all, we show that network architecture has a limited impact on the predicted mean but influences the predicted uncertainty significantly as highlighted in  \Cref{fig:bnn_network_architecture}.  We show the shallow network would be overconfident, meaning a smaller uncertainty bound, while the uncertainty would increase and gradually converge to the case of GPR shown in \Cref{fig:bnn_network_architecture}. This is reasonable because GPR is equivalent to an infinitely wide neural network with a single hidden layer \cite{Neal1995}.

\begin{figure}[h]
    \centering
    \includegraphics[width=\textwidth]{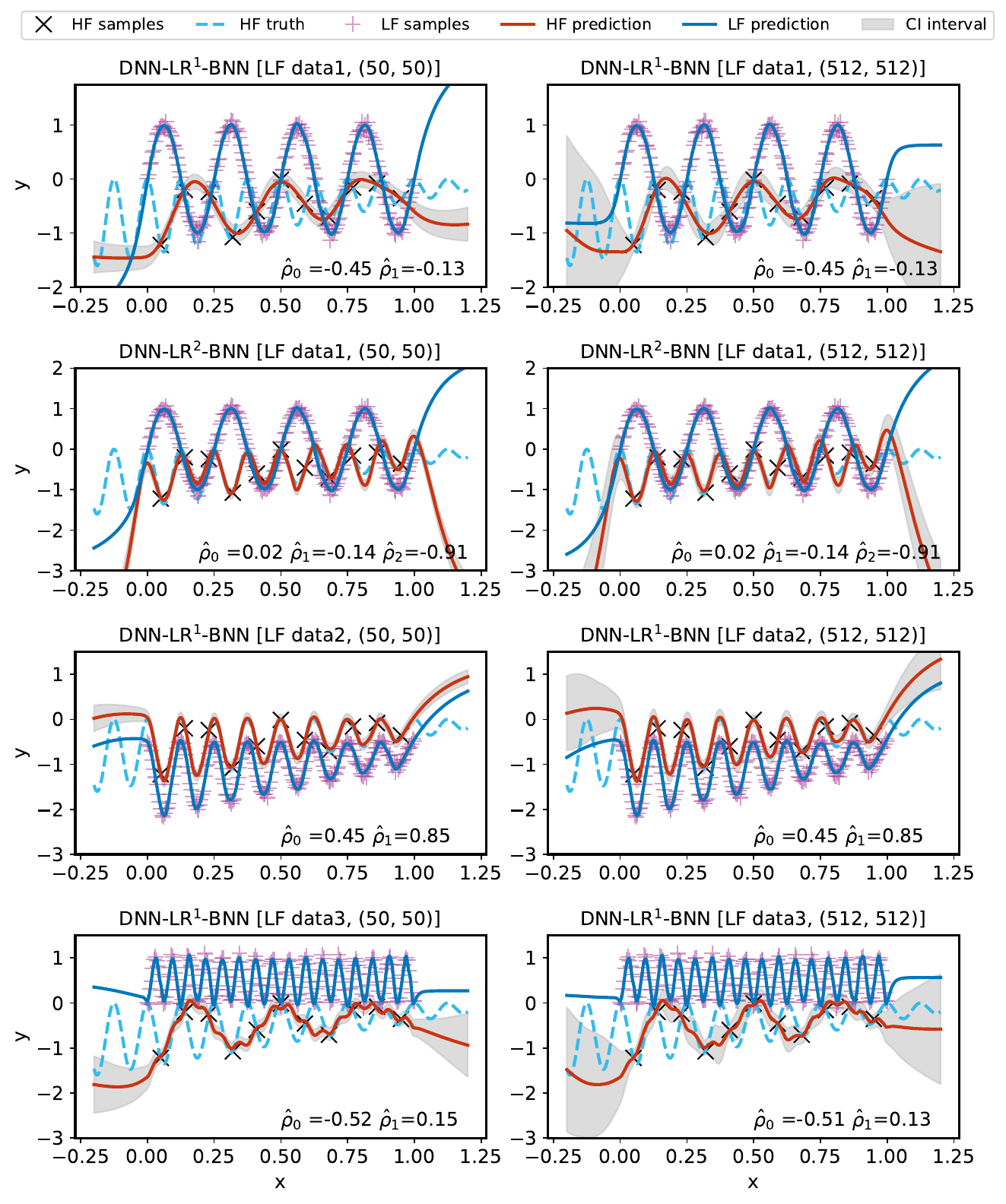}
    \caption{Comparison of DNN-LR-BNN considering two different architectures (left) using 50 neurons, and (right) using 512 neurons in each of the two hidden layers.}
    \label{fig:net_dnn_lr_bnn_influence}
\end{figure}

\begin{figure}[h]
    \centering
    \includegraphics[width=\textwidth]{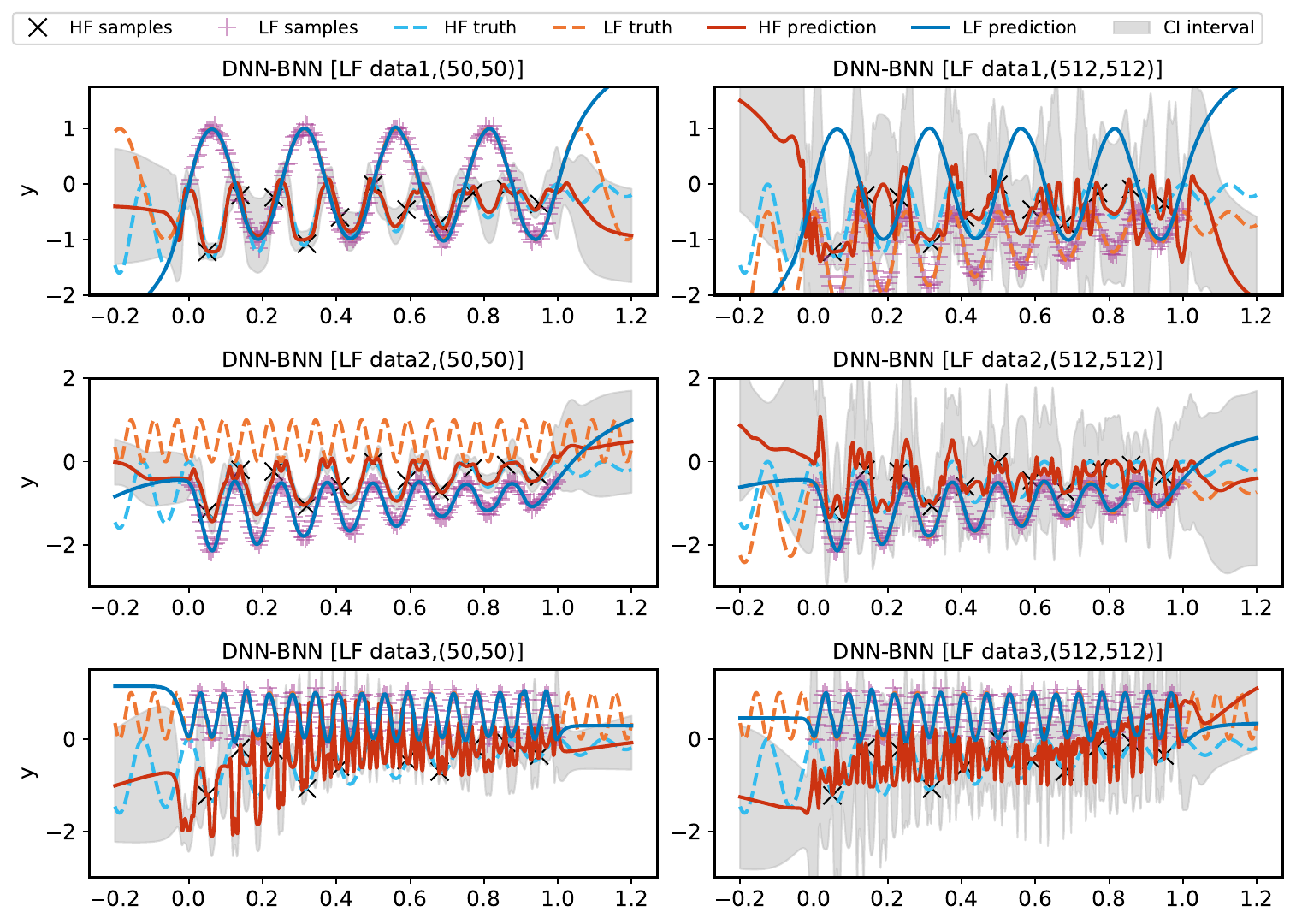}
    \caption{Comparison of DNN-BNN considering two different architectures (left) using 50 neurons, and (right) using 512 neurons in each of the two hidden layers.}
    \label{fig:net_dnn_bnn_influence}
\end{figure}

Secondly, we compare the corresponding influence of neural architecture on DNN-LR-BNN and DNN-BNN, and their results are depicted in \Cref{fig:net_dnn_lr_bnn_influence} and \ref{fig:net_dnn_bnn_influence}.  We observe the same pattern in the results of DNN-LR-BNN when enlarging the neural network architecture. Interestingly, the learned $\bm{\rho}$ is also robust to the neural network architecture choice. However, the DNN-BNN is significantly sensitive to the neural network architecture, leading to reasonable results for small networks but significant overfitting for larger networks. 

\subsubsection{Material structure-property linkages problem} 
\label{sec:additional_results_of_mat_law_prediction}

As shown in  \Cref{tab:mat_structure_property_linkage} the DNN-BNN performs poorly when using a BNN architecture with 2-layer and 256 neurons. Therefore, we conduct another experiment where we employ a smaller architecture for BNN with 2-layer and 100 neurons, and results are listed in  \Cref{tab:additional_mat_structure_property_linkage}. 

\begin{table}[h]
    \centering
    \caption{Results comparison of DNN-LR-BNN on the material structure-property linkages with 2 layer 100 neuron BNNs}
    \label{tab:additional_mat_structure_property_linkage}
    \renewcommand{\arraystretch}{1.2}
    \begin{tabular}{m{2.5cm} m{2.5cm} m{1.5cm} m{1.5cm} m{1.5cm} m{1cm} m{0.6cm} m{2cm}}
        \hline
     QoIs  &   Methods       & NRMSE    & R2 Score & TLL & $\hat{\rho}_0$& $\hat{\rho}_1$ & $\hat{\rho}_2$ \\
        \hline
         
     \multirow{3}{*}{$a_\text{eff}$ ($r$ = 0.9294)} &    
      DNN-BNN & 0.0436  & 0.9569 & -5.7493  & - & - \\
     &   DNN-LR$^1$-BNN  & 0.0603& 0.9182 & -5.7139  & 10.00 & 0.92 & - \\ 
          &   DNN-LR$^2$-BNN  & \textbf{0.0403} & \textbf{0.9633} & \textbf{-5.2613}  & 10.00 & 1.24 & -3.56 $\times 10^{-4}$ \\ 
        \hline
      \multirow{4}{*}{$b_\text{eff}$ ($r$ = 0.7681)}  
       &DNN-BNN  & 0.0115 & 0.9905  &2.9890 & - & - \\
        & DNN-LR$^1$-BNN  & \textbf{0.0111} & 0.9913 & 3.6212  & 0.10 & 0.88 & -  \\ 
        & DNN-LR$^2$-BNN  & 0.0108 & \textbf{0.9917} & 3.5549 & 0.05 & 1.17 & 0.39  \\ 
        \hline
              \multirow{4}{*}{$E_\text{eff}$ ($r$ = 0.9845)} 
              &DNN-BNN  & 0.0153& 0.9843 & \textbf{-12.1512} & - & - \\
        & DNN-LR$^1$-BNN  & \textbf{0.0107}  & \textbf{0.9923} & -12.8826  & -7.97& 1.01 & -   \\
        & DNN-LR$^2$-BNN  & 0.0130  & 0.9887 & -12.3157 & -5.24& 0.99 & 1.09 $\times 10^{-7}$  \\
        \hline

    \end{tabular}
\end{table}

We observe that the performance of DNN-BNN improves significantly compared with those in  \Cref{tab:mat_structure_property_linkage}. However, the proposed DNN-LR-BNN is still slightly better and the learned $\hat{\bm\rho}$ values are almost the same. The results demonstrate the robustness of the DNN-LR-BNN with different neuron architectures, which is an important feature for practical application.

\end{document}